\documentclass[10pt,twocolumn,letterpaper]{article}
\usepackage{bm}
\usepackage{placeins}
\usepackage{mathtools}
\usepackage{multirow}
\usepackage{color}
\usepackage[utf8]{inputenc}
\usepackage{xspace}
\usepackage{caption}
\usepackage{pifont}
\usepackage{algorithm}
\usepackage{algpseudocode}
\usepackage[table]{xcolor}
\definecolor{dancecol}{HTML}{D0E1F9}       % light blue
\definecolor{totalcol}{HTML}{DFF2D8}       % light green
\definecolor{uipcol}{HTML}{FDE2D0}         % light orange
\definecolor{gipcol}{HTML}{E3D0F9}         % light orange
\def\projname{UDP\xspace}

\DeclareMathOperator*{\argmin}{argmin}

\newcommand{\xmark}[0]{\ding{55}}
\newcommand{\cmark}[0]{\ding{51}}

\newcommand{\str}{\rlap{\textsuperscript{*}}}
\newcommand{\dagstr}{\rlap{\textsuperscript{\textdagger}}}

\usepackage{cvpr}              % To produce the CAMERA-READY version
\usepackage{makecell} % add this to your preamble
\definecolor{cvprblue}{rgb}{0.21,0.49,0.74}
\usepackage[pagebackref,breaklinks,colorlinks,allcolors=cvprblue]{hyperref}

\def\paperID{44637} % *** Enter the Paper ID here
\def\confName{CVPR}
\def\confYear{2026}

\title{Ultra Diffusion Poser: Diffusion-Based Human Motion Tracking From Sparse Inertial Sensors and Ranging-Based Between-Sensor Distances}

\author{Dominik Hollidt \quad Tommaso Bendinelli \quad Christian Holz\\
Department of Computer Science, ETH Zurich, Switzerland\\
{\tt\small \{dominik.hollidt, tommaso.bendinelli, christian.holz\}@inf.ethz.ch}
}
\hypersetup{
    pdftitle={Ultra Diffusion Poser: Diffusion-Based Human Motion Tracking From Sparse Inertial Sensors and Ranging-Based Between-Sensor Distances},
    pdfauthor={Dominik Hollidt, Tommaso Bendinelli, Christian Holz},
    pdfkeywords={sparse inertial sensing, diffusion, guidance, human pose estimation, motion capture, inertial measurement units, ultra-wideband ranging}
}
\begin{document}
\twocolumn[{%
\renewcommand\twocolumn[1][]{#1}%
\maketitle
\begin{center}
    \newcommand{\teaserwidth}{\textwidth}
% \vspace{-0.2in}
    \centerline{
    \includegraphics[width=1.0\linewidth]{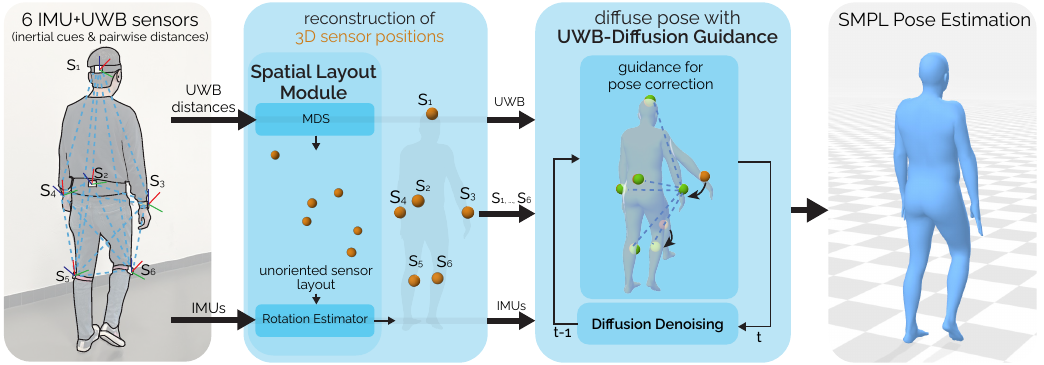}}%
  \vspace{-1mm}%
  \captionof{figure}{
Our method \projname improves wearable IMU+UWB pose estimation by extending UWB as an auxiliary feature to actively model its geometric constraints.
The Spatial Layout Module reconstructs 3D sensor positions from UWB measurements, providing a physically-informed input that conditions a diffusion model to predict SMPL poses.
UWB-Diffusion Guidance encourages alignment between predicted poses and measured distances during diffusion sampling, improving accuracy and producing consistent motions.
% Our method \projname improves wearable pose estimation from six body-worn IMUs and their pairwise UWB distances.
% \projname tightly integrates inter-sensor distances into its architecture: the Spatial Layout Module reconstructs 3D sensor positions from UWB measurements, which, together with IMU signals and UWB distances, condition a diffusion model to predict the SMPL pose.
% Finally, UWB-Diffusion Guidance encourages alignment between predicted poses and measured distances during diffusion sampling.
  }
\label{fig:teaser}
% \vspace{-3mm}
\end{center}%
}]

% \maketitle

\begin{abstract}
% \vspace*{-1mm}
Methods using inertial measurement units (IMUs) provide a wearable alternative to camera-based motion capture.
To mitigate drift from inertial signals, recent sparse inertial pose estimators integrate inter-sensor distances measured by ultra-wideband (UWB) ranging. 
So far, UWB distances have only been used as an additional input feature, ignoring the physical constraints they impose on sensor positions.
However, these distances can also be used to reconstruct the underlying 3D sensor layout, which in turn provides more informative input for pose reconstruction.
We propose Ultra Diffusion Poser, a diffusion model that explicitly models these geometric constraints.
It includes a Spatial Layout Module that analytically reconstructs the 3D sensor positions from UWB measurements.
These sensor positions are used alongside IMU signals and UWB distances as a conditioning signal during diffusion.
Still, network predictions can violate inter-sensor distance measurements.
To address this, we introduce UWB-Diffusion Guidance, which encourages alignment between predicted poses and measured distances during diffusion sampling.
Together, these contributions enable our model to achieve state-of-the-art performance, reducing joint position error by up to 22\% over prior work.
Code can be found here \url{https://github.com/eth-siplab/UltraDiffusionPoser}.
% In our evaluation, our method achieves state-of-the-art results and decreases the joint position error up to 22\% compared to previous baselines.
\end{abstract}
\section{Introduction}
\label{sec:intro}
% \begin{figure*}[ht!]
%     \centering
%     \includegraphics[width=\textwidth]{figures/main_figure_1.pdf} % Replace with your image
%     \caption{We present \projname. A system that estimates SMPL pose and translation from body-worn IMUs and their pairwise distances estimated via UWB. Through diffusion and guidance, we ensure accurate and coherent pose prediction.}
%     \label{fig:wide}
% \end{figure*}
Human pose estimation is fundamental for many applications in augmented and virtual reality and healthcare~\cite{zheng2023deep,dubey2023comprehensive}.
While numerous camera-based approaches exist to estimate human poses, they present limitations for widespread adoption. 
External cameras mounted in the environment have a fixed sensing area and suffer from occlusions when the scene is crowded~\cite{cheng2019occlusion}.
Pose estimation from wearable cameras depends on environmental factors such as lighting or background, suffers from occlusions~\cite{wang_scene-aware_2023,xu_mo2cap2_2019,tome_xr-egopose_2019, zhang2024rohm}, raises privacy concerns, and is computationally demanding~\cite{hollidt2025egosim}. 

In contrast, compact body-worn sensors such as inertial measurement units offer a minimally obtrusive alternative for human pose estimation, but accurate motion capture still requires numerous sensors.
While commercial inertial motion capture systems, such as Xsens~\cite{Xsens2024} or Noitom~\cite{Noitom2024}, achieve high accuracy with 17--19 sensors, current research is focusing on achieving the same performance with a more practical six sensor configuration ~\cite{PIP, DIP, UIP}.

While learning pipelines have improved pose accuracy, inertial approaches are inherently limited by noisy acceleration and angular velocity measurements from IMUs, which cause drifting pose estimations.
The lack of absolute measurements of IMUs, i.e., no direct observations between frames in space or across body segments, makes drift correction very hard.
As a result, sparse inertial pose estimators cannot fully resolve pose errors.

To address these limitations, recent approaches have combined IMU measurements with inter-sensor distances obtained via ultra-wideband (UWB) ranging~\cite{UIP, gip, umotion}. 
By providing non-drifting body-fixed pairwise distances between sensors, UWB adds on-body constraints that help mitigate drift and thus produce more accurate and stable pose estimates \cite{UIP, umotion}.
However, previous IMU+UWB methods have treated inter-sensor distances as auxiliary features.
Instead, enforcing them as constraints on valid sensor positions enables the recovery of the full 3D sensor layout, a more informative input than raw distance scalars.
Second, by ignoring these distance constraints during inference, existing approaches can produce poses that directly conflict with the measured distances, e.g., estimating wrist positions 70\,cm apart when UWB reports 90\,cm.

In this paper, we introduce Ultra Diffusion Poser (UDP), a diffusion-based method that explicitly models the inter-sensor distances as geometric signals, incorporating their spatial constraints directly into the model architecture.
Our approach benefits from two main novelties.
First, the Spatial Layout Module (SPL) analytically reconstructs the 3D sensor layout from just inter-sensor distances using multi-dimensional scaling (MDS).
Because the MDS reconstruction is only defined up to rotation and reflection, we introduce a learnable Rotation Estimator, trained end-to-end with the diffusion model, to recover oriented 3D sensor positions and their reflected version.
These 3D sensor positions serve as a strong conditional signal, acting as a pose prior during the diffusion denoising process.
Second, UWB-Diffusion Guidance encourages \emph{consistency} between the predicted pose and the measured inter-sensor distances, following diffusion classifier guidance~\cite{NEURIPS2021_classifier_guidance}.
During diffusion sampling, the predicted pose is continuously mapped to sensor positions via in-the-loop forward kinematics, and the alignment between predicted and measured inter-sensor distances is used to steer the denoising process toward satisfying UWB distances.
To encourage smooth predictions over long sequences, we use autoregressive diffusion inpainting~\cite{diffusionPoser,tevet2022human} that fills in the current motion, given the previous motion and the conditioning signal. 
In summary, we provide the following main contributions:

\begin{itemize}[leftmargin=*,nosep]
    \item \projname a novel fully learnable motion estimation system, that achieves state-of-the-art pose estimation from IMU+UWB measurements using an autoregressive diffusion-inpainting model. \projname explicitly models the geometric constraints imposed by inter-sensor UWB distances via the Spatial Layout Module and UWB-Diffusion Guidance.
    \item The \emph{Spatial Layout Module} reconstructs the 3D sensor positions. Using multi-dimensional scaling, the 3D layout is obtained in closed form from UWB distances and serves as a strong prior for full-body pose estimation.
    \item \emph{UWB-Diffusion Guidance} integrates in-the-loop forward kinematics into the diffusion sampling process to correct pose estimations that do not align with inter-sensor distance measurements, thereby reducing implausible or unrealistic pose estimations.
    % \item a  \textbf{autoregressive diffusion inpainting model} that accurately predicts poses from the under-constraint sparse IMU+UWB input.
    % \item a training regime that enhances pose and translation estimation through absolute translation prediction, better normalization, and the removal of forward kinematics during training.
\end{itemize}

\section{Related Work}
\label{sec:related_work}

\paragraph{Motion Capture from Sparse Inertial Sensors.}
Compared to camera-based systems, motion capture from sparse inertial sensors offers greater flexibility in placement, mobility, and robustness to occlusion. Large motion-capture datasets such as TotalCapture~\cite{trumble2017total}, DIP-IMU~\cite{DIP}, and AMASS~\cite{mahmood2019amass} have accelerated learning-based inertial pose estimation. 
These datasets enable the generation of synthetic IMU measurements by placing “virtual” sensors on mesh vertices~\cite{guzov2021human,mollyn2023imuposer,streli2023hoov,hollidt2025egosim}.
Still, synthetic data fails to fully capture real-world sensor noise and drift.

Recent methods for inertial pose estimation from six IMUs can be broadly categorized into purely learnable systems~\cite{SIP,DIP,diffusionPoser,umotion,TIP, wu2024accurate} and hybrid systems that combine neural networks with post-optimization or physics-based simulation~\cite{PIP,yi2024physical,globalpose,UIP,gip}. 
The first learning based method DIP~\cite{DIP}, introduced a recurrent neural network that learns kinematic priors from large-scale motion databases.
TIP~\cite{TIP} extends previous work by generating pose and terrain simultaneously using a transformer-based model. 
TransposePose introduces a multi-stage training process that predicts intermediate leaf joints for stabilized training. 
PIP~\cite{PIP} incorporated physical constraints (e.g., foot-floor contact) via a physics optimizer to reduce artifacts and drift. 
PNP~\cite{yi2024physical} added better physics-aware IMU simulation. 
GlobalPose~\cite{globalpose} estimates contact points into the physics simulation to reduce the flat world assumption.
In parallel, research has pushed beyond the canonical six-IMU setup by inferring the entire body from fewer sensors~\cite{ahuja2021coolmoves,jiang2023ego,zheng2023realistic,zuo2024loose}. 
While TIC~\cite{transformer_imu_calibrator} and MODA~\cite{wu2025moda} propose learning-based approaches to reduce drift, IMU-based methods remain inherently constrained by noisy inertial signals, whose integration inevitably leads to drift and degraded pose accuracy.

\paragraph{Motion Estimation with Ultra-Wideband Ranging.}
To reduce drift from inertial sensors~\cite{yao2025tof}, UIP~\cite{UIP} combines IMUs with ultra-wideband ranging to obtain direct inter-sensor distance measurements. 
Although UWB is affected by body occlusions, this fusion substantially reduces pose drift compared to IMU-only systems. 
UIP showed that passing the pairwise UWB distances through a graph convolutional network improves pose estimation. 
UMotion~\cite{umotion} proposes a Kalman filter to model and correct UWB measurement noise, but requires manual parameter tuning and cannot estimate global translation.
GIP~\cite{gip} extends UWB posing to two person pose estimation and refines their translation through an optimization framework.
While all methods make important progress in combining UWB and inertial sensing, they do not explicitly enforce pose consistency with UWB measurements or exploit the full 3D sensor layout. 
\projname uses these insights to address these limitations.
% Building on the idea of fusing UWB and IMU data, Smart Poser~\cite{devrio2023smartposer} demonstrates that even a single off-the-shelf smartwatch can leverage UWB to predict arm position.

\paragraph{Diffusion-Based Motion Estimation.}
%Particularly, diffusion-based models have shown remarkable performance in pose estimation~\cite{diffusionPoser,zhang2024rohm} and motion generation~\cite{holmquist2023diffpose,tevet2022human,jiang_motiongpt_2023,lucas_posegpt_2022}. 

Recently, methods have demonstrated the effectiveness of diffusion~\cite{ho2020denoising, song2021denoising} for motion generation \cite{tevet2022human, lucas_posegpt_2022, jiang_motiongpt_2023} and camera-based pose estimation~\cite{holmquist2023diffpose, zhang2024rohm}. 
Additionally, it has been shown that diffusion’s generative capabilities benefit underconstrained pose reconstruction from sparse trajectories~\cite{du2023avatars, yi2024estimating, ijcai2025p135}, effectively filling in missing body parts.
Similar to us, AGRoL~\cite{du2023avatars} uses a seq-to-seq approach for pose estimation from three-point tracking.
DiffusionPoser~\cite{diffusionPoser} applied a transformer-based diffusion model to inertial motion capture, enabling a greater range of potential sensor placements.
Although quantitative improvements were limited and the implementation is not publicly available, the work provides an important step toward diffusion-driven inertial pose estimation.
Building on this line of research, we introduce UWB diffusion guidance to predicted poses that are consistent with UWB distance measurements.

% \subsection{Multidimensional Scaling}
% In \projname, we explicitly reconstruct the 3D sensor layout from measured inter-sensor distances.
% A classical approach for recovering a spatial configuration of points from pairwise dissimilarities is Multidimensional Scaling (MDS)~\cite{ANOWAR2021100378, mignotte2011mds}.
% In our SPL module, we employ metric MDS (also known as classical MDS)~\cite{mead1992mds}, which assumes that the dissimilarities correspond to Euclidean distances.
% Given a pairwise distance matrix $\mathbf{D} \in \mathbb{R}^{k \times k}$, metric MDS computes an embedding of $k$ points in an $r$-dimensional Euclidean space such that the inter-point distances in the embedding approximate those in $\mathbf{D}$.
% We demonstrate that recovering 3D sensor positions from UWB distances via MDS provides a strong geometric prior that improves subsequent pose estimation.
% \input{sections/2_1_background}
\section{Method}
\label{sec:method}
We tailor \projname for the specific purpose of inertial body-pose estimation from global orientations~$\mathbf{R}$, global accelerations~$\mathbf{A}$, and pairwise inter-sensor distances~$\mathbf{D}$. 
\projname is a fully learnable autoregressive diffusion-inpainting model that integrates geometric constraints from inter-sensor distances into its architecture, first with the \textbf{Spatial Layout Module} that reconstructs 3D sensor positions using MDS from inter-sensor distances in closed-form and through \textbf{UWB-Diffusion Guidance} to align the predicted poses with the measured inter-sensor distances.

% These components, together with a normalization of our prediction windows and direct absolute translation prediction, lead to state-of-the-art results.

\subsection{Problem Statement}
\projname, similar to prior work~\cite{PIP, UIP, DIP, TIP}, aims to reconstruct human motion using $k=6$ sparse inertial sensors and their pairwise distances obtained via UWB. 

At each timestamp $t$, the inertial measurements are given by $(\mathbf{R}_t, \mathbf{A}_t)$, where $\mathbf{R}_t \in \mathbb{R}^{k \times 3 \times 3}$ represents the sensor orientations, and $\mathbf{A}_t \in \mathbb{R}^{k \times 3}$ denotes their accelerations, both expressed in a common global frame.
The pairwise distances $d_{ij}$ between sensor $i$ and $j$, are represented by matrix $\mathbf{D}_t \in \mathbb{R}^{k \times k}$.

From those inputs, \projname outputs human motion, which is parameterized using the SMPL body model $(\Theta_t, \mathbf{T}_t, \beta)$ \cite{loper2015smpl}.
$\Theta_t \in \mathbb{R}^{24 \times 3}$ defines the local joint rotations, $\mathbf{T}_t \in \mathbb{R}^3$ specifies the global body translation at time $t$, and $\beta$ is the shape vector that characterizes the individual's body shape.
% \projname also estimates the foot-ground contact probabilities $C_t\in \{0, 1\}^4$ of the heels and toes, similar to \cite{UIP, PIP, zhang2024rohm}.

Overall, \projname estimates the human motion from a sequence of N measurements, $\mathbf{R}=\{\mathbf{R}_t\}_{t=0}^N$, $\mathbf{A}=\{\mathbf{A}_t\}_{t=0}^N$, $\mathbf{D}=\{\mathbf{D}_t\}_{t=0}^N$:
\begin{equation}
    (\Theta, \mathbf{T}) = \text{\projname}(\mathbf{R}, \mathbf{A}, \mathbf{D}, \beta)
\end{equation}

\subsection{Full-Body Pose Estimation}
\begin{figure*}[ht]
    \centering
    \includegraphics[width=\linewidth]{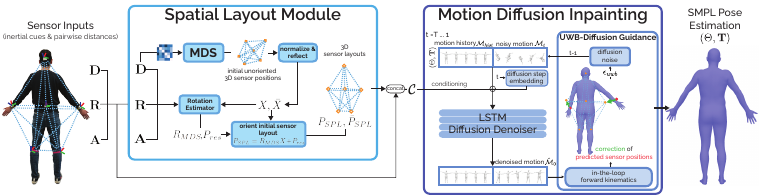}

    \caption{The Spatial Layout Module applies metric MDS to the pairwise distance matrix $\mathbf{D}$ to recover initial sensor positions, which are then oriented by a learnable Rotation Estimator.
The resulting 3D sensor layout provides a strong conditioning signal for the diffusion model.
The autoregressive diffusion inpainting model extends the previously predicted motion based on the current conditioning signal to ensure smooth motion prediction.
UWB-Diffusion Guidance steers the pose predictions to align with the measured inter-sensor distances.}
    %\caption{\projname leverages strong UWB-based distance priors. An MDS module initializes a 3D spatial layout of sensors, while diffusion combined with guidance ensures that predictions remain consistent with distance measurements.}
    \label{fig:method}
\end{figure*}
The architecture of \projname is visualized in \cref{fig:method}.
At its core, \projname is an autoregressive diffusion inpainting model that predicts the current motion given previously predicted motion and the current conditioning signal.

Instead of conditioning motion generation solely on raw IMU and UWB measurements, the Spatial Layout Module provides the 3D sensor layout that serves as a strong geometric prior during diffusion.
First, it reconstructs an initial 3D sensor layout in closed form using multidimensional scaling (MDS) based only on inter-sensor distances.
Since the MDS reconstruction is only defined up to rotation and reflection, a learnable Rotation Estimator orients both the initial MDS layout and its reflected version to produce the final oriented sensor layouts ($P_{SPL}, \bar{P}_{SPL}$).
The oriented 3D sensor layout, together with the IMU and UWB signals, form the conditioning input to the diffusion model.

To ensure that the generated poses remain consistent with measured inter-sensor distances, we introduce UWB-Diffusion Guidance.
During diffusion sampling, this mechanism uses forward kinematics (FK) to compute the predicted sensor positions from the current SMPL pose. 
These predictions are compared to the measured UWB distances, and the diffusion step is guided to correct poses that violate the constraints. 
By integrating FK directly into the sampling loop, UWB-Diffusion Guidance produces smooth and realistic motion even under noisy sensor data, whereas simple per-frame post-optimization would lead to jitter and potentially unrealistic motions~\cite{gip}.

In the following subsections, we first describe the motion representation and the conditioning signals used by \projname. We then present the components of our method: the Spatial Layout Module, the autoregressive diffusion inpainting process, and the UWB-Diffusion Guidance mechanism.

\subsubsection{Motion Representation}\label{sec:motion_representation}
To obtain the final motion estimation, \projname diffuses the motion representation conditioned on the recovered 3D sensor layout and IMU+UWB measurements.

We represent our motion sequence of length $N$ with joint orientations in 6D representation~\cite{zhouContinuityRotationRepresentations2019jun} and translation according to the SMPL model:
\begin{equation}
\mathcal{M}\in\mathbb{R}^{N\times147} = [\Theta^{6D}, \mathbf{T}]
\end{equation}

For better generalization, we normalize each motion sequence $\mathcal{M}$ such that its root joint always starts at $(0, \text{root height}, 0)$. 
This canonical normalization enables the model to predict accurate positions while mitigating representation shifts in long sequences. 

To ensure smooth and continuous motion prediction, we use autoregressive diffusion inpainting.
We prepend the $N_H$ previous predicted frames $\mathcal{M}_{hist}$ to $\mathcal{M}$, normalized such that its final frame aligns with the first frame of $\mathcal{M}$.

The diffusion model $D_\theta$ diffuses the motion representation based on a conditional signal.
This condition tensor $\bm{\mathcal{C}}$ includes the recovered 3D sensor positions $(P_{SPL}, \bar{P}_{SPL})$, sensor orientations as rotation matrices, accelerations,  inter-sensor distances, and the body shape:
\begin{equation}
\bm{\mathcal{C}} \in R^{N\times 154} = [P_{SPL}, \bar{P}_{SPL}, \mathbf{R}, \mathbf{A}, \mathbf{D}, \beta] 
\end{equation}

\subsubsection{Spatial Layout Module}\label{sec:spatial_layout_module}
The inter-sensor distances constrain the sensor positions to a fixed spatial configuration.
To recover the corresponding 3D sensor layout,  our SPL module applies metric multidimensional scaling (MDS)~\cite{mead1992mds, ANOWAR2021100378, mignotte2011mds}.
Given the pairwise distance matrix $\mathbf{D}_t \in \mathbb{R}^{k \times k}$, MDS finds the $k$ sensor points $x_i$ in 3D space whose inter-point distances best match the measured distances $d_{ij}$ by solving:
\begin{equation}
\argmin_{\mathbf{X}_{\text{MDS}} \in \mathbb{R}^{k \times 3}} \sum_{i < j} \left( | x_i - x_j |^2 - d_{ij} \right)^2.
\end{equation}
In Euclidean space, this classical MDS problem has a closed-form solution\cite{richdwilkinsonClassicalMultivariate}:
\begin{equation}
\mathbf{H} = \mathbf{I} - \frac{1}{k} \mathbf{1}\mathbf{1}^\top, \quad
\mathbf{B} = -\frac{1}{2} \mathbf{H} \mathbf{D}^{(2)} \mathbf{H},
\end{equation}
where $\mathbf{D}^{(2)}$ contains squared distances, $\mathbf{I}$ is the $k \times k$ identity matrix, and $\mathbf{1}$ is a $k$-dimensional vector of ones.

We then perform a singular value decomposition of the Gram matrix:
\begin{equation}
\mathbf{B} = \mathbf{U} \mathbf{\Sigma} \mathbf{V}^\top,
\end{equation}

and obtain the 3D sensor coordinates from the top three eigenvectors and corresponding eigenvalues:

\begin{equation}
\mathbf{X} _{\text{MDS}}= \mathbf{U}_3 \mathbf{\Sigma}_3^{1/2}.
\end{equation}

Here, $\mathbf{X}_{\text{MDS}} \in \mathbb{R}^{k \times 3}$ contains the unoriented 3D spatial positions of $k$ sensors.
This layout is defined up to translation, rotation, and reflection, and thus the sequence of layouts is temporally incoherent. 

To resolve this ambiguity, the SPL first brings the MDS layouts into a canonical, normalized frame, which is then oriented by a learned Rotation Estimator.
The initial unoriented layouts are normalized such that the pelvis is at the origin, the layout is upright (root-head aligns with y-axis) and forward facing (left wrist lies on the positive direction of the XY-plane). 
To ensure temporal consistency, frame-wise reflections are resolved whenever the current layout deviates from the previous one by more than a threshold, i.e., when \mbox{$\|X_{t-1} - X_{t}\|^2_2 > c$}. 
This yields a normalized and temporally consistent sequence of 3D sensor positions $X$.
Since the entire sequence may still be reflected, we additionally generate a mirrored version $\bar{X}$.  %$\bar{X}=R_{reflect}X$
We then retain both $X$ and $\bar{X}$ as layout candidates, one of which corresponds to the correct configuration.
Finally, a Rotation Estimator predicts the orientation of the sensor layout.
It takes the normalized layout as input and learns to rotate it into a consistent, correctly oriented configuration.
The Rotation Estimator is trained end-to-end together with the diffusion model:
\begin{equation}
 R_{MDS}, P_{res} = \text{RotEstimator}(X, \bar{X}, \mathbf{R}, \mathbf{A}, \mathbf{D}), 
\end{equation}
% The rotation matrix is predicted in 6D space \cite{zhouContinuityRotationRepresentations2019jun}
The initial MDS layout may be imperfect due to sensor noise.
Therefore, we refine it by adding an estimated residual $P_{\text{res}}$ per sensor point:
\begin{equation}
 P_{SPL} = R_{MDS}X + P_{res},\\
 \bar{P}_{SPL} = R_{MDS}\bar{X} + P_{res}, 
\end{equation}

Thus, we obtain our recovered sensor layouts $[P_{SPL}, \bar{P}_{SPL}]$ that represent the sensor positions of head, pelvis, wrists, and knees.
We use a 3-layer LSTM~\cite{lstm} for the Rotation Estimator.

\subsubsection{Motion Diffusion Inpainting}\label{sec:diff_inpainting}
\projname adopts the Denoising Diffusion Probabilistic Model (DDPM) formulation~\cite{ho2020denoising}.
At each diffusion step $t$, the denoiser $D_\theta$ predicts the clean motion $\hat{\mathcal{M}}_0$ from the noisy motion representation $\mathcal{M}_t$, conditioned on the input signal $\bm{\mathcal{C}}$ and the previously predicted motion segment $\mathcal{M}_{hist}$ of length $N_H$:

\begin{equation}
    \hat{\mathcal{M}}_0 = D_\theta(\mathcal{M}_t, \mathcal{M}_{hist}, \bm{\mathcal{C}}, t)
\end{equation}

We utilize diffusion-based inpainting~\cite{tevet2022human, diffusionPoser} to enforce temporal consistency across overlapping windows.
Specifically, the noisy motion $\mathcal{M}_t$, the motion history $\mathcal{M}_{hist}$, and the condition $\bm{\mathcal{C}}$ are projected into a shared embedding space.
The condition and noisy motion are added and appended to the motion history, and a learnable history token $h_e$ is added to mark $M_{hist}$.
The diffusion step $t$ is encoded via an MLP, added and prepended to the sequence, yielding the combined input $S$ of length $1 + N_H + N$.
We denoise $S$ using a LSTM \cite{lstm}.
%Lastly, we obtain the final denoised motion representation by projecting the segment corresponding to the current noisy input $\mathcal{M}_t$ back to the motion space.

During inference, we apply the diffusion inpainting autoregressively to predict arbitrarily long motion sequences and apply Gaussian smoothing with $\sigma=2$ for smoother results.

To train \projname we use the standard diffusion loss~\cite{ho2020denoising}:
\begin{equation} \label{eq:simple_diffusion_loss}
    \mathcal{L}_{\text{simple}} = \mathbb{E}_{q(x_t|x_0)} \left[ \| x_0 - D_\theta(x_t, t, \bm{\mathcal{C}}) \|^2_2 \right].
\end{equation}
where $q(x_t|x_0)$ denotes the DDPM forward process that gradually adds Gaussian noise to $x_0$~\cite{ho2020denoising}.
Unlike DiffusionPoser~\cite{diffusionPoser}, we train our model without the need for forward kinematics losses that significantly slow down the training. 
Instead, we use the translation and SMPL joint angle auxiliary losses:
\begin{equation}
    \mathcal{L}_{tran} = \|\mathbf{T}^{pred} - \mathbf{T}^{gt}\|_2
\end{equation}
\begin{equation}
    \mathcal{L}_{smpl} = \lvert\Theta_{6d}^{pred} - \Theta_{6d}^{gt}\rvert
\end{equation}
% \begin{equation}
%     \mathcal{L}_{contact} = \text{BinaryCrossEntropy}(C^{pred}, C^{gt})
% \end{equation}
We also utilize this velocity loss that encourages more active motion translation:
% \begin{equation}
%     \mathcal{L}_{vel} = \|(\mathbf{T}_n^{pred} - \mathbf{T}_{n-1}^{pred}) - (\mathbf{T}_n^{gt} - \mathbf{T}_{n-1}^{gt})\|_2
% \end{equation}
 \begin{equation}
      \mathcal{L}_{vel} = \left(\|\mathbf{T}_n^{pred} - \mathbf{T}_{n-1}^{pred}\|_2 - \|\mathbf{T}_n^{gt} - \mathbf{T}_{n-1}^{gt}\|_2\right)^2
  \end{equation}     

\subsubsection{UWB-Diffusion Guidance}\label{sec:classifier_guidance}
\begin{algorithm}[t]
\caption{UWB-guided conditional diffusion sampling}
\label{alg:diffusion_guidance_alg}
\begin{algorithmic}[1]
\Require Condition $\boldsymbol{\mathcal{C}}$, history $\mathcal{M}_{\mathrm{hist}}$, UWB distances $d_{ij}$
\Ensure Denoised motion $\mathcal{M}_0$
\State $\mathcal{M}_T \sim \mathcal{N}(\mathbf{0}, \mathbf{I})$
\For{$t = T$ \textbf{down to} $1$}
    \State $\hat{\mathcal{M}}_0 \gets D_\theta(\mathcal{M}_t, \mathcal{M}_{\mathrm{hist}}, \boldsymbol{\mathcal{C}}, t)$ \Comment{denoise}
    \State $\hat{d}_{ij} \gets \operatorname{FK}(\hat{\mathcal{M}}_0)$ \Comment{predicted distances}
    \State $\epsilon_{uwb} \gets \sum_{i<j} \lVert \hat{d}_{ij} - d_{ij} \rVert^2$ \Comment{UWB guidance loss}
    \State $\tilde{\mu}_t \gets \mu_t(\mathcal{M}_t, \hat{\mathcal{M}}_0) \;-\; \lambda\,\Sigma_t\,\nabla\epsilon_{uwb}$ \Comment{guidance}
    \State $\mathcal{M}_{t-1} \sim \mathcal{N}\!\big(\tilde{\mu}_t,\, \Sigma_t\big)$ \Comment{sample}
\EndFor
\State \Return $\mathcal{M}_0$
\end{algorithmic}
\end{algorithm}

Even with strong conditioning from the 3D sensor layout and UWB distances, the diffusion process can still produce motions that violate the inter-sensor distances $\mathbf{D}$.
Enforcing distances on the SMPL pose via a simple frame-wise pose optimization can lead to infeasible joint angles, unnatural poses, and jittery poses due to sensor noise.

Hence, we incorporate UWB-Diffusion guidance to steer the diffusion process toward satisfying the pairwise distance constraints, encouraging agreement with the measured inter-sensor distances $\mathbf{D}$, see \cref{alg:diffusion_guidance_alg}.

UWB-Diffusion guidance makes use of classifier guidance~\cite{NEURIPS2021_classifier_guidance} that offers a mechanism to control the diffusion sampling process. 
During inference, each noisy sample $\mathcal{M}_{t-1}$ is sampled from a Gaussian with mean $\mu_t(\mathcal{M}_{t-1}, \hat{\mathcal{M}}_0)$ that is determined by $\mathcal{M}_{t-1}$ and $\hat{\mathcal{M}}_0$, as in standard diffusion sampling~\cite{NEURIPS2021_classifier_guidance, zhang2024rohm}.
By shifting this mean in the direction of a gradient of a function $\epsilon$ that encourages desired properties, the diffusion process can be directed toward more preferred predictions (see line 6 in \cref{alg:diffusion_guidance_alg}).
In our case, the guidance term encourages consistency between the predicted pose and the measured inter-sensor distances, while retaining smooth outputs and avoiding infeasible poses.
We define the function $\epsilon$ by computing predicted sensor positions, defined by the SMPL mesh, via in-the-loop forward kinematics during sampling on $\hat{\mathcal{M}}_0$, from which the predicted distances $\hat{d}_{ij}$ are computed to match the measured distances $d_{ij}$. As UWB distances are inter-body constraints, gradients for the root translation are set to zero.
\begin{equation}
    \epsilon_{uwb} = \sum_{i < j} \| \hat{d}_{ij}(\hat{\mathcal{M}}_0) - d_{ij} \|^2
\end{equation}

% \TODO{remove}
% Following \cite{zhang2024rohm, yi2024estimating} we implement a foot skating guidance, that punishes feet sliding over the floor, via the predicted ground contact $C_{pred}$.
% \begin{equation}
%     \epsilon_{skat} = \|C_{pred} \odot \dot{FK}_{feet}(\Theta, \mathbf{T) \|}
% \end{equation}
% , where $\dot{FK}_{feet}$ represents the velocity of the 4 foot joints.

\section{Experiments}
\label{sec:experiments}
\subsection{Experimental Setup}
To ensure a fair comparison, we follow the same evaluation setup used in previous work~\cite{UIP, PIP, yi2024physical} and train on the same AMASS~\cite{mahmood2019amass} split.
We then evaluated \projname on DanceDB~\cite{AMASS_DanceDB}, an AMASS split held out during training,  TotalCaptureReal~\cite{trumble2017total} and the test split of DIP-IMU~\cite{DIP}, using the DIP-IMU training split. 

Finally, to demonstrate real-world applicability, we also evaluated our model on two real-world datasets where raw IMU and UWB are recorded during acquisition: UIP-DB~\cite{UIP} and the recent GIP-DB~\cite{gip}. 
These datasets are particularly challenging due to higher and more realistic IMU drift and UWB noise.
For these datasets, we report the performance without, i.e. zero-shot, and with fine-tuning. 
Further details regarding the training and evaluation protocol are available in the supplementary material (\cref{seq:training_details}). \\
We evaluated our method using the following metrics:

\begin{description}[leftmargin=!,labelsep=0.2em]

    \setlength{\itemsep}{-2pt}  % Space between items
    \setlength{\parsep}{0pt}   % Space between paragraphs within an item
    \setlength{\topsep}{-2pt}    % Space before and after the list
    \setlength{\partopsep}{0pt} % Extra space when list starts at a paragraph break
  \item[\textbf{SIP} ($^\circ$)] global angle error of shoulders and hips.
  \item[\textbf{GAE} ($^\circ$)] mean global angle error of all joints.
  \item[\textbf{JPE} (cm)] root-aligned joint position error.
  % \item[\textbf{TE@2m}/\textbf{TE@5m} (m)] accumulated root translation error.
  \item[\textbf{Jitter} (m/s$^{3}$)] mean magnitude of the third derivative (jerk) of global joint positions.
\end{description} 

We compare \projname against state-of-the-art methods, including the  IMU+UWB approaches UIP~\cite{UIP}, UMotion~\cite{umotion} and GIP~\cite{gip}, as well as the IMU-only methods PNP~\cite{yi2024physical}, GlobalPose~\cite{globalpose}, DynaIP~\cite{zhang2024dynamic}, PIP~\cite{PIP}, and TIP~\cite{TIP}. 
Note that, to ensure fairness, we report the numbers from the original papers and replicate results using our pipeline whenever possible.

\subsection{Results}
We report quantitative results for DIP-IMU in \cref{tab:res_dip} and for DanceDB, TotalCapture, UIP-DB, and GIP-DB in \cref{tab:sip_gae_jpe_jitter}, with detailed interpretation in the following subsection. 
Next, we analyze \projname's sensitivity to UWB noise (\cref{sec:noise_analysis}).
Finally, multiple ablation studies are presented to assess the effectiveness of our modules (\cref{sec:ablation}).
Additional results and experiments are available in the appendix.

\subsubsection{Pose Estimation Evaluation}
UDP achieves state-of-the-art results across most evaluation settings. 
UDP achieves the lowest joint position error in all evaluation settings, with up to 22\% improvement, except for UIP-DB without fine-tuning. 
The improvement over IMU-only methods is up to 35\% showing the benefit of distance measurements.
Notably, UDP produces consistently smooth motion with minimal jitter across all datasets without the need for a physics optimizer as in UIP, PIP, or PNP. \\
% \projname produces superior results, especially when the UWB data is accurate, as in DIP-IMU, DanceDB ,and TotalCapture.
On DIP-IMU, DanceDB, and TotalCapture the SIP error is consistently the lowest, highlighting the accurate estimation of the limbs, which impacts visual quality the most.
The GA error achieved by \projname is either the lowest or competitive, being only slightly surpassed by DynaIP on DIP-IMU and by GlobalPose on TotalCapture.
This demonstrates that modeling geometric constraints from inter-sensor distances improves \projname’s pose estimation.

\projname also generalizes well to real-world datasets. 
On GIP-DB, a newer real-world dataset with modern UWB sensors and lower noise levels, \projname achieves state-of-the-art results in both zero-shot and fine-tuned settings. 
In contrast, it only struggles when both UWB and IMU signals are highly noisy, as seen on UIP-DB, where IMU drift reaches 3.21°/min and UWB errors 17.5 cm~\cite{UIP}. 
In this case, \projname achieves SOTA performance only after fine-tuning, which allows it to adapt to the dataset’s noise characteristics, while remaining competitive otherwise. 
We posit that, as \projname relies more heavily on UWB, it is more susceptible to UWB sensing errors; we investigate this further in \cref{sec:noise_analysis}.

\begin{table}[ht]
    \centering
    \caption{Results on DIP-IMU. "*" denotes the results we reproduced, because they were not provided in the original paper. "\textdagger" denotes results that we reproduced, because the paper used a different evaluation framework. GT Jitter is 1.830.}
    \setlength{\tabcolsep}{3pt} 
    \scalebox{0.9}{
    \begin{tabular}{lcccccccc} % ground truth jitter is (1.829)
        \toprule
         Model & UWB & SIP (°) $\downarrow$ & GAE (°) $\downarrow$ & JPE (cm) $\downarrow$  & Jitter \\
         \midrule
         TIP & \xmark & 17.07  & 10.51\str & 5.82 & 0.882 \\
         PIP & \xmark & 15.02  & 8.78\str & 5.12 & 0.240 \\ % GAE and pos error taken from PNP
         PNP & \xmark &13.71 & 8.75 & 4.98 & 0.260 \\
         GlobalPose & \xmark & 13.55 & 8.47 & 4.65  & 0.260 \\
         % \multicolumn{4}{c}{\TD{figure out what to do with dynaip}} \\
         DynaIP& \xmark & 14.41 & \textbf{7.12}\str & 5.03 & - \\
         % UIP & 13.200 & - & 5.050 & 0.240\\
         % UIP (Ying) & 11.503 & 8.231 & 4.028 & 0.140 \\
         % \projname old & \textbf{10.41} & \textbf{8.20} & \textbf{3.44} & \textbf{0.163} \\
         UMotion & \cmark & 15.05\dagstr & 10.41\dagstr & {4.38}\dagstr & {0.216}\dagstr \\
         UIP & \cmark & {13.20} & 8.23\str & 5.05 & 0.240 \\
         \projname (ours) & \cmark & \textbf{10.39} & {8.19} & \textbf{3.42} & \textbf{0.125} \\
         \bottomrule
    \end{tabular}
    }

    \label{tab:res_dip}
\end{table}

\begin{table*}[ht]
    \centering
    \caption{Combined colored results on DanceDB (blue), TotalCapture (green), and UIP-DB (orange). “*” denotes values we reproduced because they were not reported in the original papers. "\textdagger" denotes results that we reproduced because the paper used a different evaluation framework.
    Ground Truth Jitter: DanceDB: 2.09, TotalCapture: 0.465, UIP-DB: 0.850, GIP-DB: 0.113. \textbf{Bold} indicates best.
    }
    \scalebox{0.77}{
    \begin{tabular}{llccccc}
        \toprule
        Dataset       & Model                   & UWB & SIP (°) $\downarrow$ & GAE (°) $\downarrow$ & JPE (cm) $\downarrow$ & Jitter \\
        \midrule
        \rowcolor{dancecol}
        DanceDB       & TIP                     & \xmark & 18.70                     & 11.80\str              & 8.50                  & 1.438    \\
        \rowcolor{dancecol}
        DanceDB       & PIP                     & \xmark & 20.00                     & 19.17\str             & 8.87                  & 0.661\str \\
        \rowcolor{dancecol} 
        DanceDB       & PNP                     & \xmark & 16.52\str & 11.45\str & 7.18\str & 0.747\str \\
        \rowcolor{dancecol} 
        DanceDB       & UMotion                 & \cmark & 11.93\dagstr & 10.61\dagstr & 5.19\dagstr & 1.471\dagstr \\   
        \rowcolor{dancecol}
        DanceDB       & UIP                     & \cmark & 15.28                     & 10.98\str             & 7.45                  & 0.430    \\
        \rowcolor{dancecol}
        DanceDB       & \projname (ours)        & \cmark & \textbf{11.79}            & \textbf{9.91}         & \textbf{4.67}         & \textbf{0.175}    \\
        \addlinespace
        \rowcolor{totalcol}
        TotalCapture  & TIP                     & \xmark & 11.36                     & 12.30                 & 5.15                  & 0.751   \\
        \rowcolor{totalcol}
        TotalCapture  & PIP                     & \xmark & 12.93                     & 12.04                 & 5.61                  & 0.204   \\
        \rowcolor{totalcol}
        TotalCapture  & PNP                     & \xmark & 10.89                     & 10.45                 & 4.74                  & 0.260   \\
        \rowcolor{totalcol} 
        TotalCapture  & GlobalPose              & \xmark & 9.81 & \textbf{9.99} & 4.25 & 0.350 \\
        \rowcolor{totalcol}
        TotalCapture  & DiffusionPoser          & \xmark & –                         & 14.40                 & 6.10                  & –              \\
        \rowcolor{totalcol} 
        TotalCapture  & UMotion                 & \cmark & 10.58\dagstr & 11.36\dagstr & 4.83\dagstr & 0.215\dagstr \\
        \rowcolor{totalcol}
        TotalCapture  & UIP                     & \cmark & 10.70                     & 11.43\str             & 5.11                  & 0.206   \\
        \rowcolor{totalcol}
        TotalCapture  & \projname (ours)        & \cmark & \textbf{8.95}             & 10.19       & \textbf{3.76}         & \textbf{0.124}   \\
        \addlinespace
        \rowcolor{uipcol}
        UIP-DB        & TIP                     & \xmark & 33.01                     & -              & 14.82                 & 1.860   \\
        \rowcolor{uipcol}
        UIP-DB        & PIP                     & \xmark & 30.47                     & 29.33\str             & 13.62                 & 1.570   \\
        \rowcolor{uipcol} 
        UIP-DB        & UMotion                 & \cmark & 31.61\dagstr & 27.92\dagstr & 13.19\dagstr & 0.054\dagstr \\
        \rowcolor{uipcol} 
        UIP-DB        & UIP                     & \cmark & \textbf{24.12}            & 27.59\str             & \textbf{10.65}        & \textbf{0.050}   \\
        \rowcolor{uipcol}
        % UIP-DB        & \projname (ours)        & \cmark & 25.84                     & \textbf{26.29}        & 11.92                 & 0.069   \\ % with normal guidynce
        UIP-DB        & \projname (ours)        & \cmark & 24.95                    & \textbf{25.92}        & 11.72                 & 0.071   \\ % with 5e1 guidance
        \midrule
        \rowcolor{uipcol} 
        UIP-DB        & UMotion finetuned & \cmark & 24.37\dagstr & 25.69\dagstr & 11.13\dagstr & \textbf{0.045}\dagstr \\
        \rowcolor{uipcol}
        UIP-DB        & UIP finetuned           & \cmark & 23.85                     & 25.34\str             & 10.65                 & 0.080   \\
        \rowcolor{uipcol}
        % UIP-DB        & \projname (ours) finetuned     & \cmark & \textbf{21.23}            & \textbf{23.30}        & \textbf{9.52}         & 0.074   \\ % with normal guidance
        UIP-DB        & \projname (ours) finetuned     & \cmark & \textbf{19.24}            & \textbf{21.81}        & \textbf{9.04}         & 0.074   \\ % with 5e1 guidance
        % \midrule % <-- keep separator between UIP-DB and GIP-DB
        \addlinespace
        \rowcolor{gipcol} 
        GIP-DB        & UIP                     & \cmark & 30.18 & 26.16 & 10.88 & 0.224\str \\
        \rowcolor{gipcol} 
        GIP-DB        & UMotion                 & \cmark & 26.57\dagstr & 23.43\dagstr & 9.52\dagstr & 0.128\dagstr \\
        \rowcolor{gipcol} 
        GIP-DB        & GIP                     & \cmark & 27.77 & 23.34 & 9.45 & 0.706\str \\
        \rowcolor{gipcol} 
        % GIP-DB        & \projname (ours)        & \cmark & \textbf{25.99} & \textbf{22.30} & \textbf{9.04} & \textbf{0.129} \\ % normal guidance
        GIP-DB        & \projname (ours)        & \cmark & \textbf{25.74} & \textbf{22.23} & \textbf{8.86} & \textbf{0.125} \\ % 5e1 guidance
        \midrule
        \rowcolor{gipcol} 
        GIP-DB        & UMotion finetuned & \cmark & 25.67\dagstr & 23.23\dagstr & 9.34\dagstr & 0.132\dagstr \\
        \rowcolor{gipcol} 
        GIP-DB        & GIP finetuned           & \cmark & 18.04 & 17.57 & 8.70 & 0.608\str \\
        \rowcolor{gipcol} 
        % GIP-DB        & \projname (ours) finetuned & \cmark & \textbf{15.35} & \textbf{15.12} & \textbf{7.40} & \textbf{0.108} \\ % normal guidance
        GIP-DB        & \projname (ours) finetuned & \cmark & \textbf{15.33} & \textbf{14.34} & \textbf{6.68} & \textbf{0.092} \\ % 5E1 guidance
        \bottomrule
    \end{tabular}
    }
    \label{tab:sip_gae_jpe_jitter}
\end{table*}

%\begin{table*}[]
%    \centering
%    \scalebox{0.9}{
%    \begin{tabular}{lccccccc}
%        \toprule
%         Metric & SIP Error (°) $\downarrow$ & GAE (°) $\downarrow$ & JPE (cm) $\downarrow$ & TE @ 2m  $\downarrow$ & TE @5m $\downarrow$ & Jitter (0.850) \\
%         \midrule
%         TIP & 33.01  & 57.63\str & 14.82 & 0.61\str & 0.82\str & 1.860 \\ % todo fix GAE and translation
%         PIP & 30.47  & 29.33\str & 13.62 & \textbf{0.31}\str & \textbf{0.42}\str & 1.570 \\
         % UIP (paper) & 24.12 & - & 10.65 & & & - \\
%         UIP & \textbf{24.12} & 27.59\str & \textbf{10.65} & 0.34\str & 0.45\str & 0.050 \\
         % \projname (old)& 25.80 & \textbf{26.73} & 11.91 & 0.41 & 0.79 & 0.066  \\
%         \projname & 25.84 & \textbf{26.29} & 11.92 & 0.62 & 0.88 & 0.069  \\
  %       \midrule
 %        UIP finetuned & 23.85 & 25.34\str & 10.65 & 0.31\str & \textbf{0.41}\str & 0.080 \\
         % \projname finetuned (old)  & \textbf{21.04} & \textbf{23.40} & \textbf{9.52} & \textbf{0.30} & 0.61 & 0.071  \\
   %      \projname finetuned  & \textbf{21.23} & \textbf{23.30} & \textbf{9.52} & \textbf{0.50} & 0.83 & 0.074  \\
    %     \bottomrule
    %\end{tabular}
    %}
%    \caption{Results on UIP-DB. * denotes the results we reproduced, because they were not provided in the original paper.}
%    \label{tab:res_uip_db}
% \end{table*}

\subsubsection{Qualitative Results}\label{sec:qualitative_results}
\begin{figure}[htb]
    \centering
    \includegraphics[width=1\linewidth]{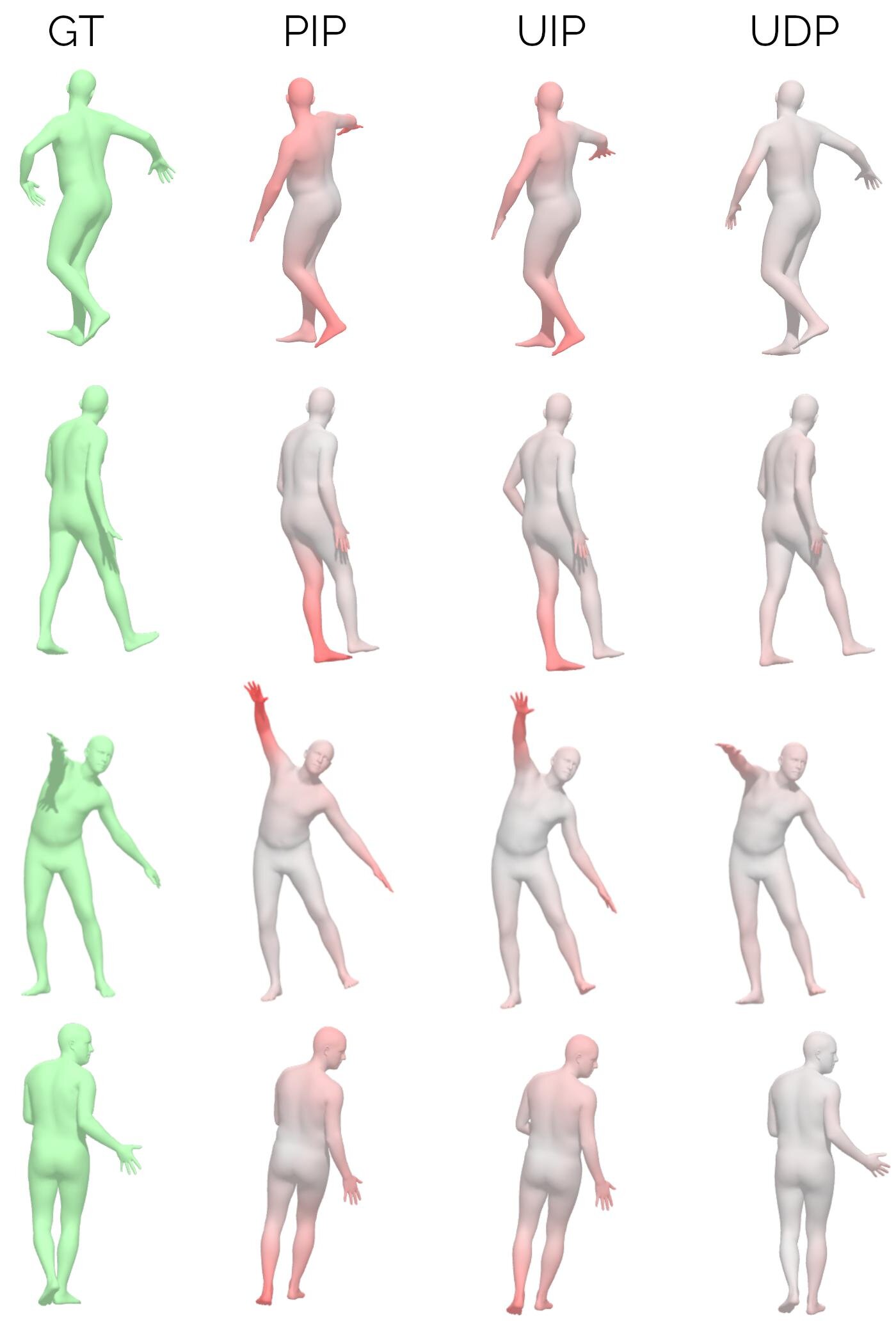}
    \caption{Qualitative Results of \projname from DanceDB and TotalCapture. A brighter red indicates a higher position error. \projname consistently produces accurate poses.}
    \label{fig:qualitative_results}
\end{figure}

We present extensive qualitative results in \cref{fig:qualitative_results} and the supplementary material.

As seen in the figure, UDP excels in the accurate estimation of limb positions. 
This is due to its strong integration of inter-sensor distances, which impose strong constraints on feasible positions.
For example, in row one of \cref{fig:qualitative_results}, the right arm is extended slightly backward and the left foot is raised slightly higher, resulting in a more accurate pose prediction compared to PIP and UIP.
Notably, the better pose estimations of \projname are most noticeable during fast motions such as fast arm movements, where other methods tend to lag behind.
Thanks to the UWB-Diffusion Guidance, \projname remains robust, producing more accurate leg estimates, unlike physics-optimized methods such as PIP, PNP, and UIP, that struggle with complex dance sequences and yield implausible foot placements, see second row in \cref{fig:qualitative_results}.

\subsubsection{Sensitivity Towards Noise}\label{sec:noise_analysis}
While \projname is the best method on all other datasets, it achieves only competitive performance in the zero-shot setting on the UIP-DB dataset, where UWB measurement errors exceed 17 cm.
Here, UIP surpasses it in SIP and joint position error.
We attribute this to the strong influence of noisy UWB signals, which are tightly integrated into \projname's architecture.
To analyze this, we compare both methods on UIP-DB by progressively blending noisy UWB measurements with ground-truth distances (\cref{tab:uwb_blending_results}).
While UIP performs slightly better under raw, noisy conditions, UDP achieves up to 1.4 cm lower JPE and 4.59° lower GAE once measurement noise is reduced.
This trend aligns with \projname's SOTA results on GIP-DB, a newer real-world dataset with more modern, less noisy UWB sensors, demonstrating both its real-world applicability and the importance of modern UWB-radios with a lower noise profile.

\begin{table}[ht]
    \centering
    \caption{Improvement of \projname~over UIP across different UWB noise levels, non-finetuned. The UWB distances are linearly interpolated between real and ground-truth measurements.}
    \begin{tabular}{lcc}
        \toprule
        \multirow{2}{*}{UWB-Noise Level} & \multicolumn{2}{c}{Improvement over UIP} \\
         & GAE (°) & JPE (cm) \\
        \midrule
        100\% (Real noise) & 1.67 & -1.07 \\
        75\%  & 3.19 & 0.71 \\
        50\%  & 3.76 & 0.91 \\
        25\%  & 4.12 & 1.14 \\
        0\% (Perfect distances) & 4.59 & 1.40 \\
        \bottomrule
    \end{tabular}
    \label{tab:uwb_blending_results}
    % \vspace{-5pt}
\end{table}

\subsection{Ablation Studies}\label{sec:ablation}
To validate the contributions of our two main modules, we perform ablation studies on TotalCapture.
\begin{table}[ht]
    \centering
    \setlength{\tabcolsep}{3pt}
    \caption{Ablation Studies on TotalCapture. }
    \resizebox{\columnwidth}{!}{
    \begin{tabular}{lccc}
    \toprule
         Ablation & SIP (°) $\downarrow$ & GAE (°) $\downarrow$ & JPE (cm) $\downarrow$ \\
         \midrule
         \projname w/o SPL \& UWB Guidance & 10.22 & 10.79 & 4.39 \\
         \projname w/o SPL & 9.58 & 10.45 & 4.03 \\
         \projname w/o UWB Guidance & 9.46 & 10.46 & 4.05 \\
          \projname w/o RotEstimator & 9.11 & 10.45 & 3.92 \\
          \projname RotEstimator w/o $P_{res}$ & 9.01 & 10.22 & 3.79 \\
         \projname & \textbf{8.95} & \textbf{10.19} & \textbf{3.76} \\
         \bottomrule
    \end{tabular}
    }
    \label{tab:ablation_rest}
\end{table}

\noindent\textbf{Spatial Layout Module.} Disabling the SPL module increases all metrics, with SIP error rising by 7\%, highlighting its impact on overall pose accuracy, see \cref{tab:ablation_rest}.
The SPL benefits further from the Rotation Estimator, which orients sensor layouts.
Disabling it and replacing the oriented sensor layouts $P_{SPL}, \bar{P}_{SPL}$ with the normalized MDS layouts $X,\bar{X}$ increases all errors.
This shows that reconstructing the 3D sensor layout via SPL yields better pose estimates than just using inter-sensor distances as auxiliary inputs.

\noindent\textbf{UWB-Diffusion Guidance.} UWB-Diffusion guidance improves local pose estimates, reducing SIP error by 5\% and JPE by 7\%. 
Limb segments benefit most from the corrective effect of diffusion guidance.
Disabling both the SPL and UWB-Diffusion Guidance further deteriorates performance, with JPE increasing by 17\%, showing their complementary effect.
Because UWB-Diffusion Guidance applies corrections during the sampling process, the predicted motions remain smooth.
This highlights the benefit of leveraging geometric constraints during diffusion sampling by correcting poses that violate UWB measurements.

\section{Discussion and Limitations}
\label{sec:discussion}
We show that \projname excels at local pose estimation thanks to its strong integration of inter-sensor distances. 
However, this makes it sensitive to very noisy UWB measurements, see \cref{sec:noise_analysis}, and the UWB-Guidance coefficient should be adjusted to the expected noise level.
Future work could improve robustness by modeling sensor uncertainty, which can be incorporated into the SPL module or guidance loss.

While \projname is fully data-driven and does not require manual tuning of post-optimization problems, Kalman filter hyperparameters, as in UMotion, or modeling of physics optimizers, it also does not benefit from the constraints these approaches enforce.
For tasks requiring zero-foot skating, ground penetration avoidance, or strict physical consistency, especially when floor or environment information is available, physics optimizers still have an advantage.
Future work could investigate refining \projname’s predictions using existing physics optimizers or integrating novel physics-based guidance terms into the diffusion framework.

\projname’s main contributions focus on improving local pose through on-body constraints.
This could be extended to inter-person constraints, similar to GIP~\cite{gip}, by utilizing inter-person UWB distances.
Combining UDP’s smooth, UWB-aligned motion predictions with the more jitter-prone inter-person optimizations from GIP has the potential for accurate and globally consistent motion estimation.
These ideas could further be extended to incorporate environment-fixed UWB beacons, enabling multi-person and environment-aware pose refinement.

Finally, similar to prior work~\cite{PIP, UIP, umotion, globalpose, yi2024physical}, \projname is restricted to a fixed six-sensor setup. Extending support to arbitrary sensor layouts could further increase its real-world applicability and flexibility.

% Across all datasets, it produces smooth and accurate motions without relying on a physics optimizer, and it is fully learnable; requiring neither Kalman filter hyperparameter tuning (as in UMotion) nor post-processing adjustments.
% With the availability of UWB measurements, our novel SPL module and UWB-Diffusion guidance reduce joint position errors by 22-32\% IMU+UWB/IMU based methods, respectively.
% Yet, due to its strong integration of inter-sensor distances \projname is very sensitive to wrong UWB measurements. 
% Still \projname provides SOTA results on real-world datasets GIP-DB and UIP-DB.
% Future work should consider improving the robustness towards measurement noise, such as by integrating an uncertainty-aware SPL module or guidance loss.
% Despite improving translation prediction, \projname still suffers from translational drift that would require external drift correction, which could be an important capability to be further extended for real-world applications.
% Lastly, while using shape parameters during training, we assume the mean body shape during evaluation. 
% Future work could better estimate body shape from the existing sensor distance data, which potentially could further decrease pose errors during inference.
\section{Conclusion}
\label{sec:conclusion}
In this work, we introduced \projname, a novel full-body pose estimation approach utilizing sparse inertial sensing combined with UWB-based distance measurements.
While prior methods used UWB distances only as additional features, \projname explicitly models the geometric constraints they impose on sensor positions.
It integrates three key components: (1) a novel Spatial Layout Module that estimates 3D sensor positions, (2) an autoregressive diffusion inpainting model, and (3) UWB-Diffusion Guidance that encourages pose predictions to be aligned with measured distances during diffusion sampling.
We show that explicitly using the Spatial Layout Module and UWB-Diffusion Guidance leads to more accurate and physically plausible poses compared to treating UWB distances as auxiliary input features.
In total, our experiments showed that \projname improves pose estimation by up to 22\%, outperforming both its direct IMU+UWB baselines and IMU based posers.
We hope these results highlight the potential of modeling the geometric constraints provided by novel UWB sensors explicitly to produce advanced models for more generalized and reliable wearable-based pose estimation solutions.

%We demonstrated that \projname improves pose estimation by up to 43\% and translation improvements by up to 70\%. 
%Extensive ablation studies show the benefits of our model choices.

\paragraph{Acknowledgements}
We thank Yi-Chi Liao, Jiaxi Jiang, and all other people who provided valuable feedback.

{
    \small
    \bibliographystyle{ieeenat_fullname}
    \bibliography{main}
}


\appendix

\newpage
\clearpage
\FloatBarrier

\section{Supplementary}
\begin{figure*}[ht]
    \centering
    \includegraphics[width=0.9\linewidth]{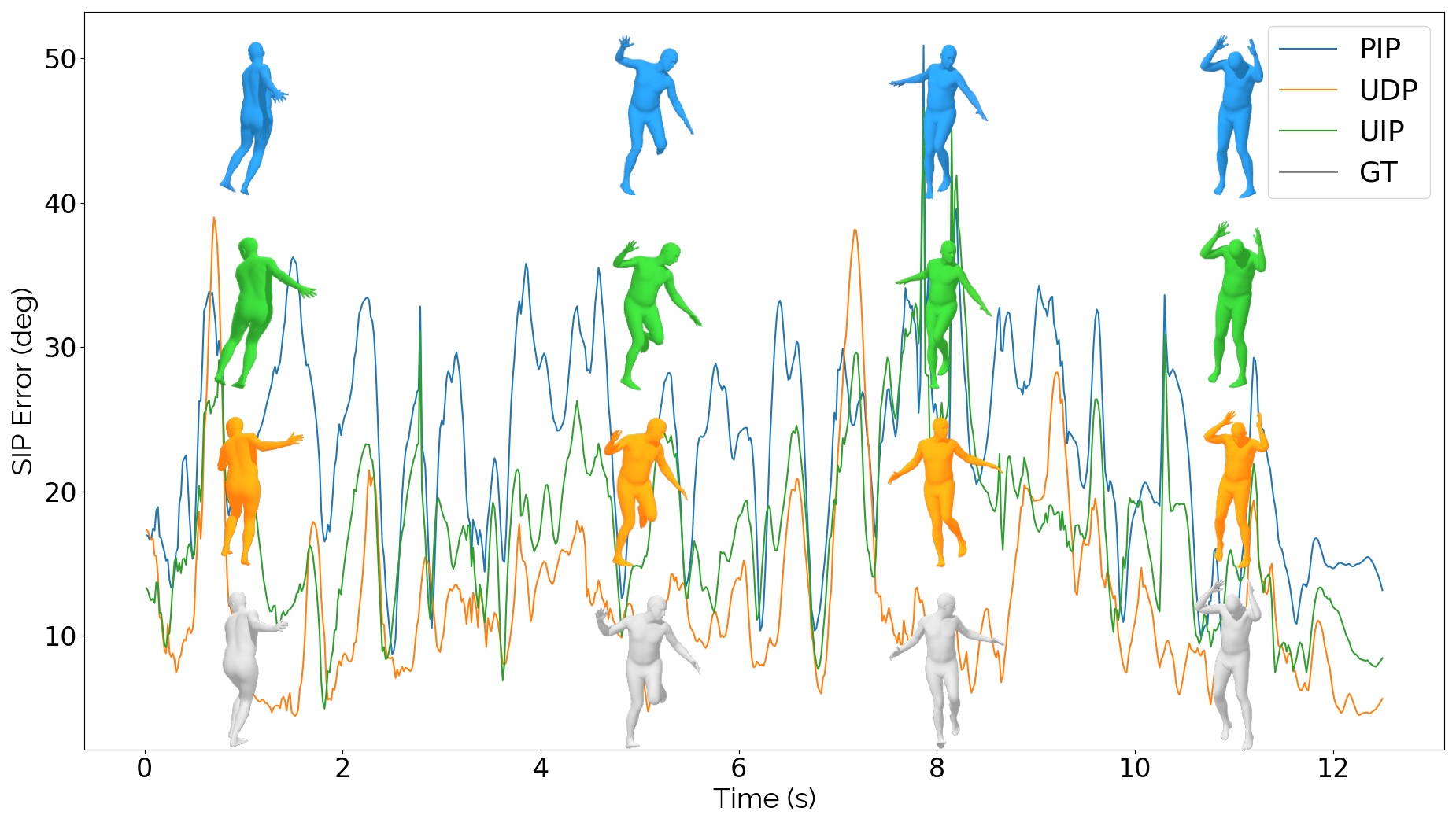}
    \caption{Qualitative results over a sequence of DanceDB. \projname has generally lower SIP error compared to previous methods. }
    \label{fig:sip_qual}
\end{figure*}
\subsection{Training and Evaluation Details}\label{seq:training_details}
In the following section, we provide additional implementation and evaluation details.

The losses described in \cref{sec:diff_inpainting} are weighted by the following weights: $\lambda_{\text{simple}} = 5$, $\lambda_{\text{tran}} = 5$, $\lambda_{\text{smpl}} = 1$, and $\lambda_{\text{vel}} = 0.3$. 
The velocity loss is activated after 10 epochs, once the prediction has stabilized. 

\projname uses an LSTM with a hidden state size of 512 with 2 layers and a dropout rate of 0.4 for the denoiser network. 
The motion history length is set to $N_H=30$, and the motion window size is $N=170$ frames at 60 FPS during training. 
The model is trained for 150 epochs with a weight decay of $1\mathrm{e}{-5}$, an initial learning rate of $1\mathrm{e}{-4}$, and step decay with a factor of $\gamma=0.33$ every 50 epochs. 
We use a square root diffusion noise schedule.

The rotation estimator uses a hidden state dimension of 512, 3 layers, and a dropout rate of 0.2. We enable UWB-Diffusion Guidance at diffusion step $t=20$, with $\lambda = 500$ and a weaker $\lambda = 50$ for the GIP-DB and UIP-DB, where UWB-signals are less reliable.
Overall, the model contains 11.3M parameters, and training takes approximately 5 hours on an RTX 4090 GPU with a batch size of 256.

Following prior work~\cite{PIP, TIP, UIP}, the model is initially trained on the synthetic AMASS dataset with DanceDB held out and directly evaluated on TotalCaptureReal and DanceDB. 
This is the complete list of Datasets used during training: HumanEva, MPI\_HDM05, SFU, MPI\_mosh, Transitions\_mocap, SSM\_synced, CMU, TotalCapture, Eyes\_Japan\_Dataset, KIT, BMLmovi, EKUT, TCD\_handMocap, ACCAD, BioMotionLab\_NTroje, BMLhandball, MPI\_Limits, DFaust\_67.

DIP-IMU and TotalCapture provide IMU signals coming from an XSense suit~\cite{Xsens2024}.
For TotalCapture, we use the DIP sensor calibration.
All AMASS training data and DanceDB use synthesized IMU data, obtained via finite differences on the motion data following~\cite{PIP, UIP}.
UWB measurements are synthesized from the motion ground truth for AMASS training data, DanceDB, TotalCapture, and DIP-IMU.
UIP-DB~\cite{UIP} and GIP-DB~\cite{gip} obtain all IMU data through off-the-shelf, low-cost commercial IMU and UWB sensors, and we use these real-world IMU signals and UWB measurements directly.
For DIP-IMU, the model is fine-tuned on the training split for 30 epochs and then evaluated on the test split. 
As UIP-DB, and GIP-DB present a more challenging scenario with lots of real-world sensor noise, we train this model with noise augmentations.
We introduce Gaussian noise with $\sigma_{\text{acc}} = 0.6$ to the acceleration measurements and apply random rotation augmentation with Gaussian noise of $\sigma_{\text{ori}} = 0.15$, and add a random Gaussian bias of $\sigma_{\text{ori-bias}} = 0.054$ and a stronger weight decay of $4e-5$.
For UWB distance noise, we add Gaussian noise with the standard deviation based on the reported mean UWB distance errors from UIP-DB \cite{UIP} scaled by 1.5.
The model is pretrained for 50 epochs using 200 diffusion steps to improve robustness to noise.
For fine-tuning, we take the model pretrained on purely synthetic data and train it for 5 epochs on the training split of either UIP-DB or GIP-DB (depending on which dataset is being evaluated).
We then evaluate the model on the corresponding test split.

For consistency, we report results from the original authors or from previously published works. If the required metrics are not available, we reproduce the results ourselves by utilizing the provided weights. 
We retrain the network if no weights are provided for that evaluation.
Since PIP does not provide training code, we implemented our own training pipeline and trained the model from scratch, ensuring that no test data from DanceDB leaked into the training process.
PNP~\cite{yi2024physical} does not provide data processing code and provide the reported results on TotalCapture and DIP.
DynaIP does not predict translation; hence, we only report the results of the translation-free DIP dataset.
UMotion uses a slightly different evaluation scheme; we used the authors' provided model weights and reran the results.
During inference, we disable the use of ground-truth distances for updating the Kalman filter covariances, see \href{https://github.com/kk9six/umotion/blob/6213c32433c34200d6d7de262eda36d84cedbafd/src/main_uip.py\#L70}{here}. 
Instead, we assume an 8 cm standard deviation for the UWB measurements and set the measurement covariance accordingly.
Additionally, we report results in two settings:
First, the zero-shot results reported on UIP-DB and GIP-DB use the raw UWB distances.
Second, the finetuned results that clean the raw UWB distances using the training split, as seen \href{https://github.com/kk9six/umotion/blob/6213c32433c34200d6d7de262eda36d84cedbafd/src/data/uip.py\#L53}{here in the authors code}.
The reported results are obtained following the same evaluation framework as previous work~\cite{PIP, UIP, yi2024physical}.
Evaluation metrics follow the conventions established in prior work~\cite{UIP, PIP, yi2024physical}, excluding end joints that cannot be reliably estimated from sparse sensor input—specifically, the wrists, hands, root, toes, and ankles.

% ===== TABLE 2: TE @ 2m and TE @ 5m =====

\subsection{Additional Evaluations}

\begin{figure*}[ht]
    \centering
    \includegraphics[width=0.69\linewidth]{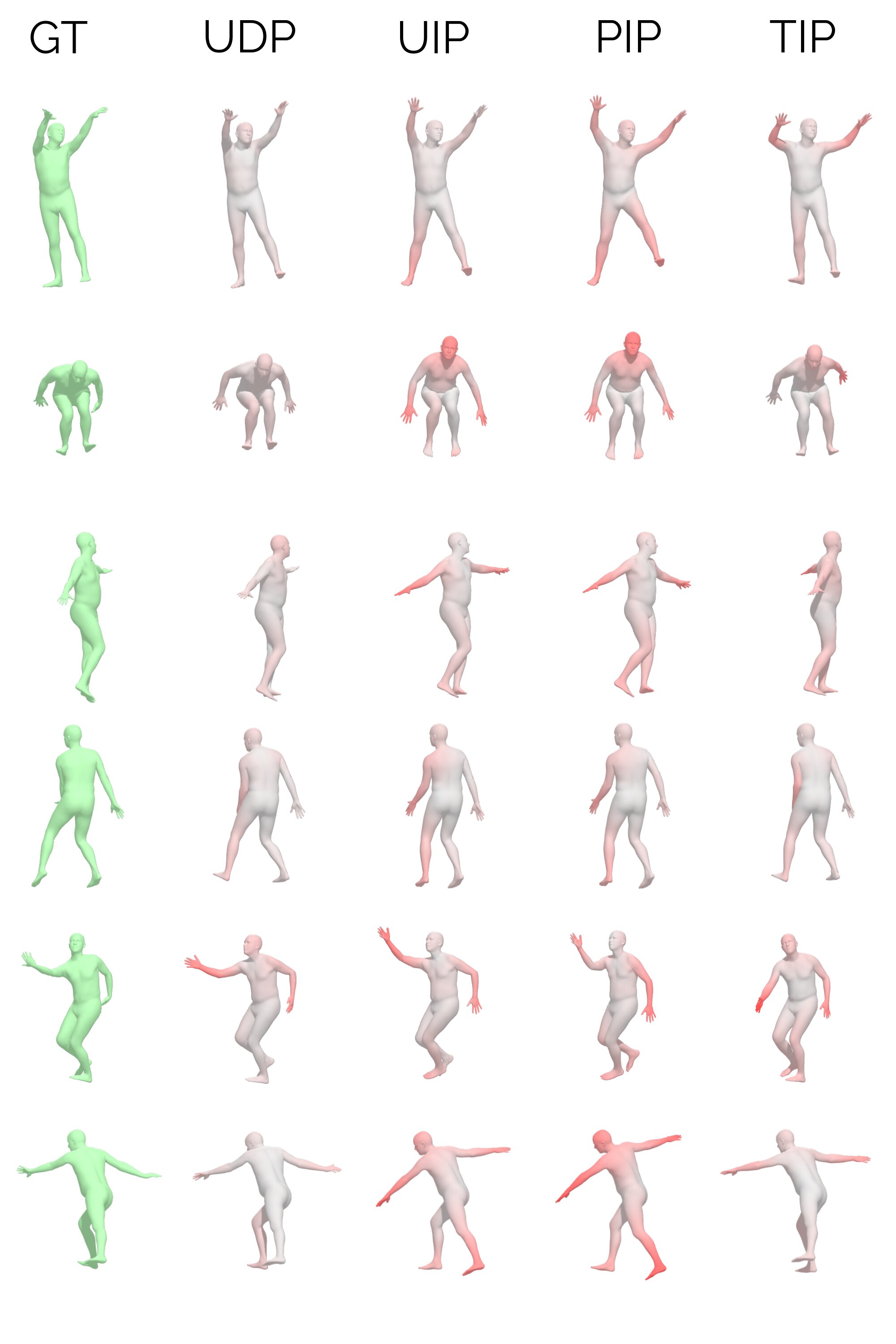}
    \caption{Qualitative results on DanceDB. \projname often estimates fast motions more accurately than other methods.}
    \label{fig:additional_qual_results_dancedb}
\end{figure*}

\begin{figure*}[ht]
    \centering
    \includegraphics[width=0.7\linewidth]{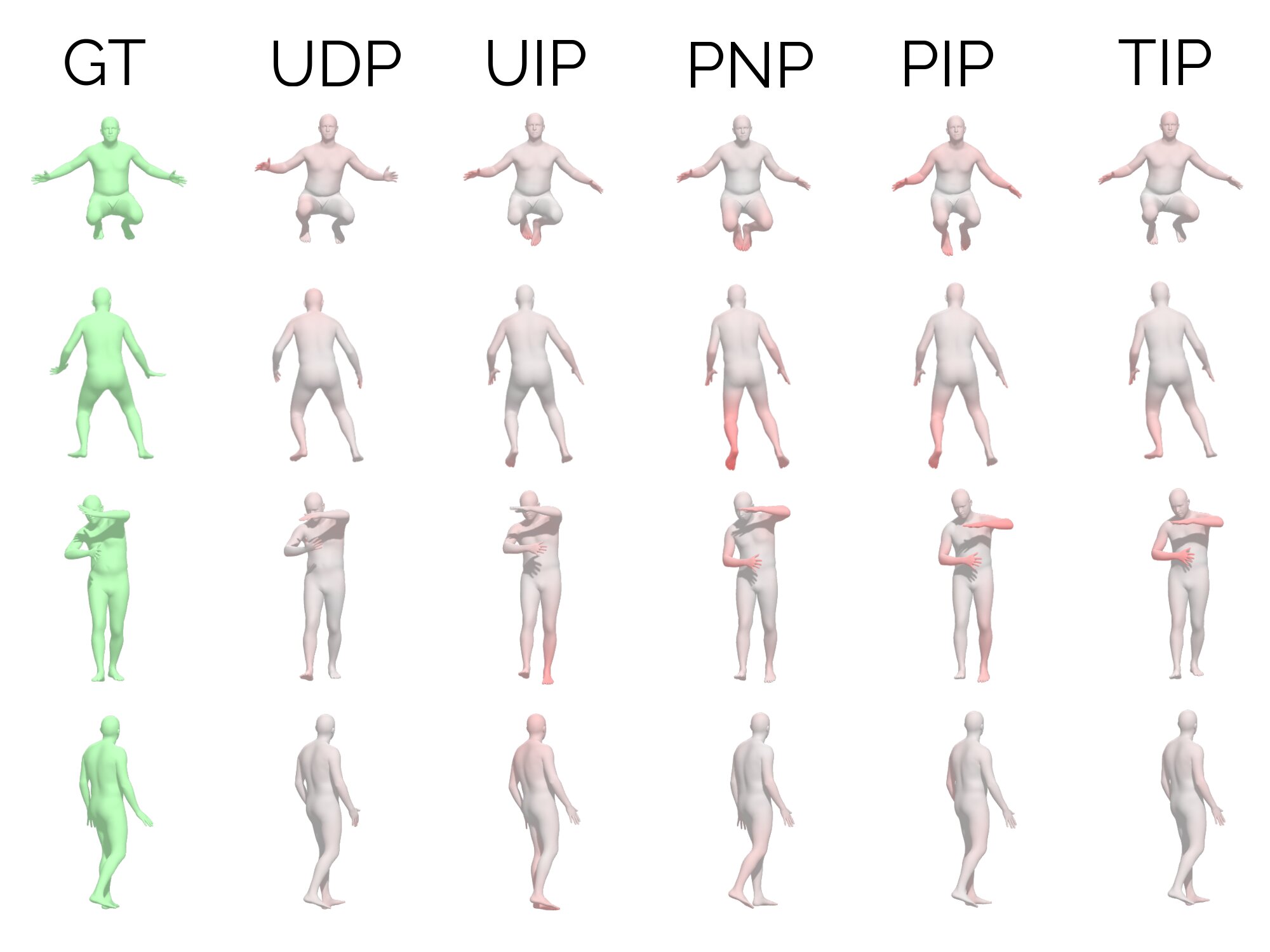}
    \caption{Qualitative results on TotalCapture. \projname often estimates limb positions more accurately than others.}
    \label{fig:additional_qual_results_total_capture}
\end{figure*}

We provide additional qualitative comparisons in the accompanying video and in \Cref{fig:additional_qual_results_dancedb,fig:additional_qual_results_total_capture}.
As shown, \projname achieves better foot and arm placement than prior work, particularly in squatting motions, where it better estimates foot positioning while maintaining precise torso alignment.

Conversely, methods relying on post-optimization physics simulations (e.g., PIP, UIP, PNP) often struggle with highly dynamic motions, such as dance routines involving rapid twirls or intentional foot sliding (see the bottom row of \cref{fig:additional_qual_results_dancedb}).
In these scenarios, the physics optimizer may be a bottleneck due to rigid constraint modeling. 
Specifically, when the network predicts a foot contact, the optimizer enforces strict zero-velocity or high-friction constraints. 
Consequently, intentional sliding motions may lead to wrong motion estimation when the model inaccurately predicts foot contacts.
In contrast, \projname, being purely data-driven, is not bound by contact enforcement, friction models, or hyperparameter choices within the physics model and avoids such interference, allowing for the accurate reconstruction of complex maneuvers.

While physics optimizers remain a compelling alternative for applications where strict physical plausibility (e.g., non-penetration) takes precedence over tracking fidelity, our results demonstrate that \projname yields smooth, accurate motion, particularly in limb placement, while benefiting from the simplicity of a purely data-driven approach.

\subsubsection{Additional Ablation Studies}
The following sections extend the analysis in the main paper by providing detailed ablation studies regarding inference dynamics, noise scheduling, temporal lookahead, and architectural alternatives.

\subsubsection{Sampling Steps, Inference and Realtime}
\projname operates in real-time and significantly improves inference speed over UIP.
Unlike UIP, it avoids the post-processing physics optimizer, which formulates refinement as a Quadratic Programming (QP) problem.
This QP solver acts as a computational bottleneck, as it relies on serial CPU execution and cannot benefit from GPU acceleration.
In contrast, \projname utilizes purely neural-based inference that is fully accelerated on the GPU.

\cref{tab:inference} shows that \projname achieves 3.6 times faster inference compared to UIP when running on an RTX 4090.
By using DDIM~\cite{song2021denoising}, we can reduce sampling steps to just 5. This increases inference speed by a factor of 14 compared to DDPM sampling with 50 steps, while the SIP error increases by only 4.8\%.
We observe that the pose prediction quality stabilizes quickly: the transition from 50 steps (DDPM) to 25 steps (DDIM) induces the largest drop in quality, while further reducing the steps to 5 has a negligible impact.
In comparison to UIP, \projname with 5 sampling steps enables 50 times faster inference while still delivering superior output quality.
% Nevertheless, all evaluated methods run fast enough to enable real-time motion estimation on modern hardware.

\begin{table}[ht]
    \centering
    \caption{Evaluation of the total inference time (TT) on the TotalCapture dataset (approx. 2,466,s duration) and the mean inference time per frame (MTPF). For 60FPS real-time inference, 16,ms MTPF or lower is required. Reducing the number of diffusion steps reduces pose estimation accuracy slightly but increases inference speed drastically.}
    \resizebox{\linewidth}{!}{
    \begin{tabular}{lccccc}
        \toprule
         Metric & SIP (°)  $\downarrow$ & JPE  (cm) $\downarrow$ & TT (s)  $\downarrow$ & MTPF (ms)  $\downarrow$ \\
         \midrule
         TIP & 11.36     &  5.15    & 568  & 3.938 \\
         UIP & 10.70 & 5.11 & 1,342 & 7.545 \\
         UDP DDIM 5 & 9.38 & 4.01 & \textbf{26} & \textbf{0.151} \\
         UDP DDIM 10 & 9.34 & 3.99 & 46 & 0.265 \\
         UDP DDIM 25 & 9.30 & 3.98 & 110 & 0.627 \\
         UDP DDPM 50 & \textbf{8.95} & \textbf{3.76} & 371 & 2.117 \\
         \bottomrule
    \end{tabular}
    }
    
    \label{tab:inference}
\end{table}

\subsubsection{Module Ablation}
We showed that the SPL module and UWB-Diffusion Guidance improve pose estimation under ideal sensing conditions (TotalCapture). 
\cref{tab:ablation_gip_db} demonstrates that both components also improve accuracy under noisy sensor conditions. 
Notably, adapting the UWB guidance coefficient $\lambda$ to the sensor noise level enables improved performance, as noisy distance measurements with strong guidance can induce wrong limb predictions.

\begin{table}[h!]
    \centering
    % Tighten vertical space
    \caption{GIP-DB ablation on GIP-DB finetuned: SPL and UWB-Guidance improve metrics under real-world noise.}
    \resizebox{\columnwidth}{!}{
    \begin{tabular}{lccc}
        \toprule
         Metric & SIP (°) $\downarrow$ & GAE (°) $\downarrow$ & JPE (cm) $\downarrow$ \\
         \midrule
         \projname w/o SPL & 16.03 & 15.09 & 7.08 \\
         \projname w/o UWB Guidance & 16.30 & 14.86 & 6.90 \\
         \projname $\lambda=500$ & 15.35 & 15.12 & 7.40 \\ % normal guidance
         \projname & \textbf{15.33} & \textbf{14.34} & \textbf{6.68} \\
         
         \bottomrule
    \end{tabular}
    }
    \label{tab:ablation_gip_db}
\end{table}

% \begin{table}[h!]
%     \centering
%     % Tighten vertical space
%     \caption{GIP-DB ablation: SPL and UWB-Guidance improve metrics under real-world noise. $\Delta>0$ is worse performance.}
%     \resizebox{\columnwidth}{!}{
%     \begin{tabular}{lccc}
%         \toprule
%          UDP & $\Delta$ SIP (°) & $\Delta$ GAE (°) & $\Delta$ JPE (cm) \\
%          \midrule
%          w/o Spatial Layout Module & +0.37 & +1.43 & +1.09 \\
%          w/o UWB Guidance & +0.96 & +0.52 & +0.22 \\
%          \bottomrule
%     \end{tabular}
%     }
%     \label{tab:ablation_delta}
% \end{table}

\subsubsection{Noise Schedules}
\begin{figure*}[ht]
    \centering
    \includegraphics[width=0.8\linewidth]{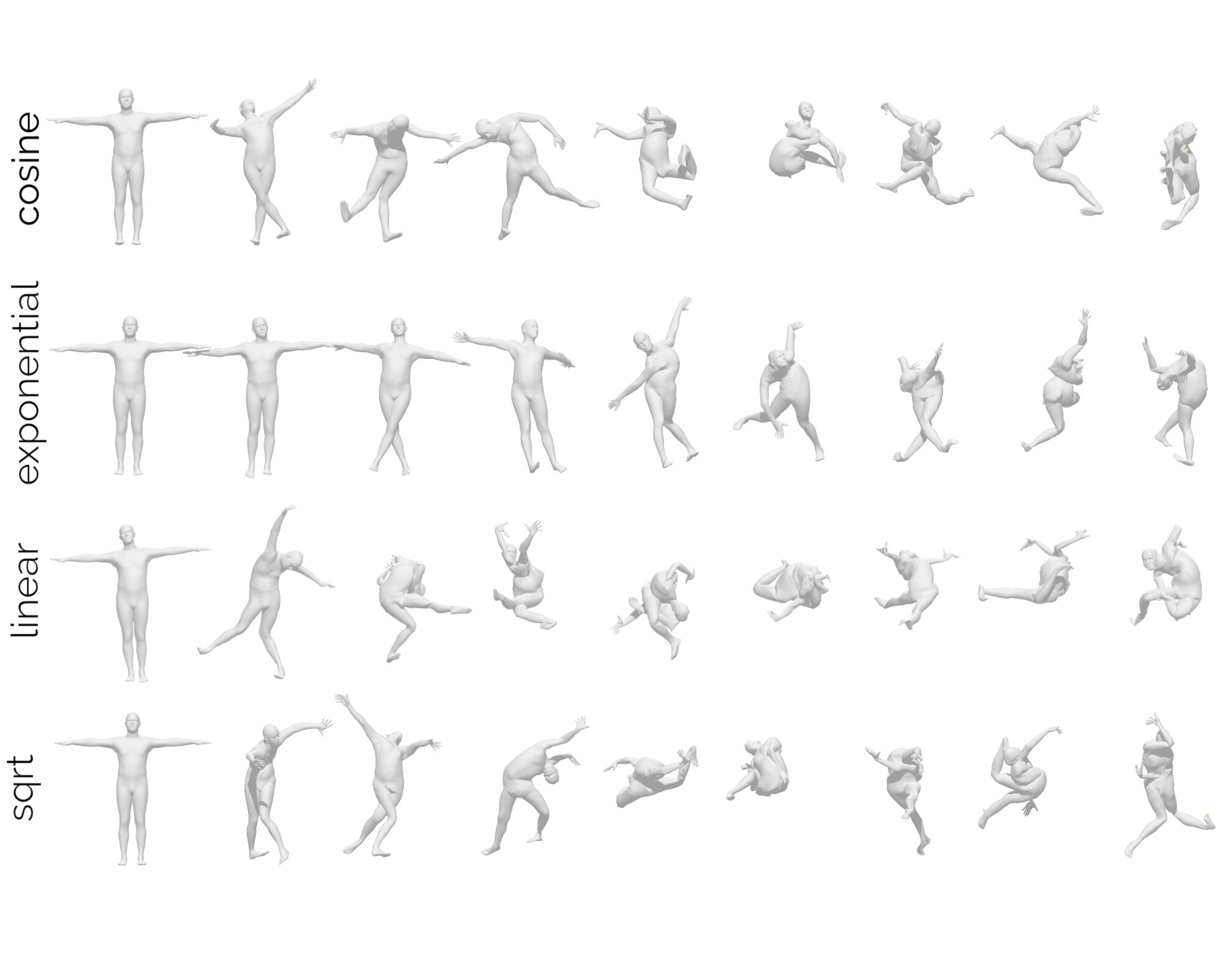}
    \caption{Noise schedules of different schedules visualized on $q(x_t|x_0)$ in equal steps where the maximum step is $T=50$.}
    \label{fig:noise}
\end{figure*}

Additionally, we evaluated various noise schedules, as diffusion models are often sensitive to the specific noise distribution of the data.
We hypothesized that schedules adding noise more gradually (e.g., cosine or exponential) would be beneficial, given that error accumulation along the kinematic chain causes big joint errors with even minimal joint angle noise.
As illustrated in \cref{fig:noise}, aggressive schedules like the square root schedule can rapidly degrade the pose structure, leading to visually distorted samples even at early noise steps.
Counter-intuitively, the more aggressive square root schedule led to the best performance.
As shown in \cref{tab:res_noise_schedule}, the square root schedule consistently outperforms all other tested schedules across all metrics.

\begin{table}[ht]
    \centering
    \begin{tabular}{lccccc}
         \toprule
         Metric &  SIP (°) $\downarrow$ & GAE (°) $\downarrow$ & JPE (cm) $\downarrow$\\
         \midrule
         cosine & 9.18 & 10.25 & 3.87 \\ 
         exponential & 9.29 & 10.42 & 4.00  \\ 
         linear & 9.46 & 10.48 & 4.09 \\
         sqrt & \textbf{8.95} & \textbf{10.19} & \textbf{3.76}  \\
         \bottomrule
    \end{tabular}
    \caption{\projname trained with different noise schedules.}
    \label{tab:res_noise_schedule}
\end{table}

\begin{table*}[ht]
    \centering
    \setlength{\tabcolsep}{3pt} 
        \caption{Ablation Study results on TotalCapture.}
    \scalebox{0.99}{
    \begin{tabular}{lccccccc}
    \toprule
         Metric & SIP Error (°) $\downarrow$ & GAE (°) $\downarrow$ & JPE (cm) $\downarrow$ & TE @ 2m  $\downarrow$ & TE @5m $\downarrow$ & Jitter (0.465) \\
         \midrule
         % \projname w/o UWB distance inputs & 14.18 & 12.63 & 6.44 & 0.20 & 0.48 & 0.128 \\
         % \projname w/ Transformer old & 12.66 & 13.48 & 5.40 & 0.38 & 0.82 & 0.175 \\
         \projname w/ Transformer & 10.70 & 12.10 & 4.79 & 0.35 & 0.55 & 0.115 \\
         % \projname w/o MDS old& 9.58 & 10.45 & 4.03 & 0.20  & 0.48 & 0.124 \\
         % \projname w/o SPL & 9.58 & 10.45 & 4.03 & 0.34  & 0.53 & 0.125 \\
         % \projname w/ velocity old & 9.19 & 10.31 & 3.92 & 0.17 & 0.38 & 0.116 \\
          \projname w/ velocity representation & 9.30 & 10.35 & 3.95 & 0.28 & 0.49 & 0.123 \\
         % \projname w/o UWB Guidance OLD & 9.29 & 10.40 &  3.97 & 0.13 & 0.28 & 0.118 \\
          % \projname w/o UWB Guidance & 9.46 & 10.46 &  4.05 & \textbf{0.20} & \textbf{0.32} & 0.126 \\
         % \projname w/ FK loss old & 8.94 & 10.30 & 3.86 & \textbf{0.12} & \textbf{0.27} & 0.116 \\
         \projname w/ FK loss & 9.01 & 10.29 & 3.89 & \textbf{0.20} & \textbf{0.32} & 0.124 \\
         % \projname old & \textbf{8.81} & \textbf{10.15} & \textbf{3.70} & 0.13 & 0.29 & 0.116  \\
         \projname (ours) & \textbf{8.95} & \textbf{10.19} & \textbf{3.76} & \textbf{0.20} & 0.33 & 0.124 \\
         \bottomrule
    \end{tabular}
    }
    \label{tab:ablation_full}
\end{table*}

\subsubsection{Shaped Evaluation}
UDP predicts SMPL joint angles and root translation, while the SMPL shape parameters $\beta$ serve as optional conditioning input when available.
Our main evaluation sets $\beta=\mathbf{0}$ (mean shape) for the model input and evaluation.
$\beta$ can be user-provided or estimated from UWB distances (as in UMotion~\cite{umotion}).
We show that utilizing the known $\beta$ reduces UDP's errors (Tab.~\ref{tab:shaped_eval}), improving robustness across diverse body proportions. When using ground-truth shape parameters both as model input and for evaluation, \projname achieves consistent gains across all metrics, most notably a 10\% reduction in SIP error. Since shape influences both inter-sensor distances and inertial dynamics, providing $\beta$ further enhances motion estimation accuracy.
\begin{table}[h!]
    \centering
    \caption{Evaluation DanceDB. Shape improves motion estimation.}
    \resizebox{\columnwidth}{!}{
    \begin{tabular}{lccc}
        \toprule
         Model & SIP (°) $\downarrow$ & GAE (°) $\downarrow$ & JPE (cm) $\downarrow$ \\
         \midrule
         UDP w/ known shape & \textbf{10.68} & \textbf{9.35} & \textbf{4.30} \\
         UDP w/o known shape & 11.79 & 9.91 & 4.67 \\
         \bottomrule
    \end{tabular}
    }
    \label{tab:shaped_eval}
\end{table}

\subsubsection{No Lookahead}
We evaluate the model's performance in an online setting by restricting the evaluation to the last frame of the input window (Last Frame Only - LFO).
In this scenario, the model relies solely on past observations without access to future context.
As shown in \cref{tab:res_lookahead_tc}, removing future context results in slightly higher errors, i.e., approximately 2.7\% in SIP, while the global angle error remains largely unaffected.
Additionally, reducing the context window from 3s to 1s increases JPE by 3\%, with a marginal impact on angle errors.
Crucially, even in this restricted online setting with no lookahead, \projname continues to outperform baseline methods.

\begin{table}[ht]
    \centering
    \scalebox{0.95}{
    \begin{tabular}{ccccc}
         \toprule
         LFO & W length &  SIP (°) $\downarrow$ & GAE (°) $\downarrow$ & JPE (cm) $\downarrow$ \\
         \midrule
         \checkmark & 1\,s & 9.36 & 10.45 & 4.14 \\ 
          & 1\,s & 9.15 & 10.49 & 3.84\\
         \checkmark & 3\,s & 9.19 & \textbf{10.11} & 4.01 \\ 
         & 3\,s & \textbf{8.95} & 10.19 & \textbf{3.76}  \\
         \bottomrule
    \end{tabular}
    }
    \caption{Ablation on lookahead and window size. 'LFO' denotes evaluating only the last frame of the window.}
    \label{tab:res_lookahead_tc}
\end{table}

% \begin{figure}
%     \centering
%     \includegraphics[width=0.8\linewidth]{figures/cumulative_translation_errors_TotalCapture.pdf}
%     \caption{The cumulative translation error on TotalCapture. All models perform similarly. The still}
%     \label{fig:translation_error_total_capture}
% \end{figure}

\subsubsection{Removal of FK loss}
Unlike previous methods~\cite{diffusionPoser, zhang2024dynamic}, we do not utilize a forward kinematics loss during training.
Using the FK loss increases the training time from approximately 5h to 14h.
Furthermore, as shown in \cref{tab:ablation_full}, the forward kinematics loss does not compromise performance; in fact, it yields a marginal reduction in pose estimation accuracy.
We attribute this to unstable gradients arising from the kinematic chain: despite similar loss values, the FK loss produces up to 10× higher gradient norms and 1000× larger peak gradients with respect to SMPL joint angles compared to a direct L1 loss. This indicates that direct supervision of local joint angles is sufficient.

\subsubsection{Model Architecture}
Previous works on pose estimation~\cite{TIP, jiang2022avatarposer, diffusionPoser} and pose generation~\cite{tevet2022human} have employed transformers.
However, our experiments show that a transformer encoder~\cite{vaswani2017attention} performs significantly worse as a denoising model, with a 27\% increase in joint position error, see \cref{tab:ablation_full}.
We implemented a 4-layer transformer encoder with the same token dimension as \projname, using sinusoidal position encodings.
Overall, our results indicate that the local neighborhood priors captured by LSTMs are more effective than those learned by transformers, given the dataset size used in our experiments.

\subsection{Translation Results}

\begin{figure}[ht]
    \centering
    \includegraphics[width=1\linewidth]{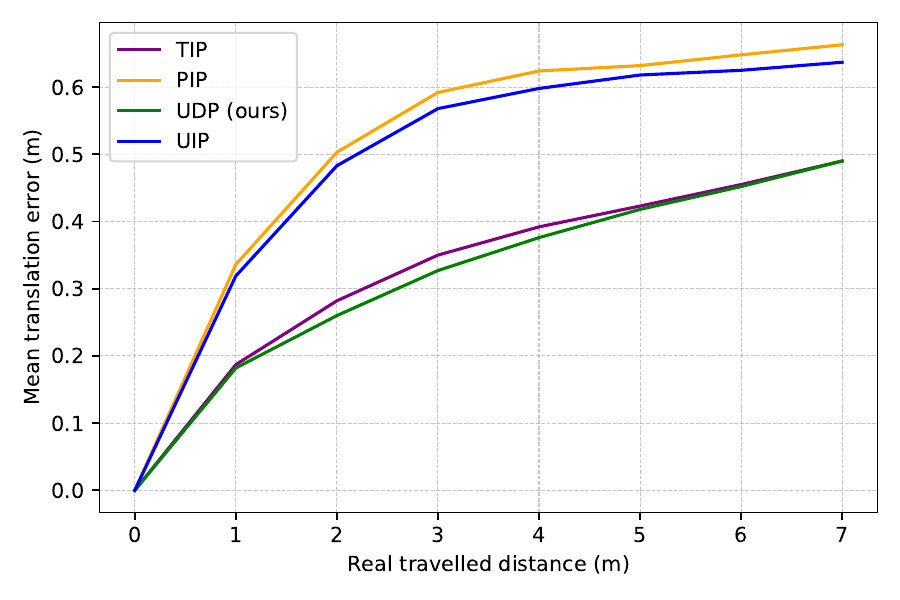}
    \caption{The cumulative translation error over segment of $d$ meters on DanceDB. Purely learned methods UDP and TIP, tend to perform better on complicated dance moves.}
    \label{fig:translation_error_dance_db}
\end{figure}

\begin{figure}[ht]
    \centering
    \includegraphics[width=0.9\linewidth]{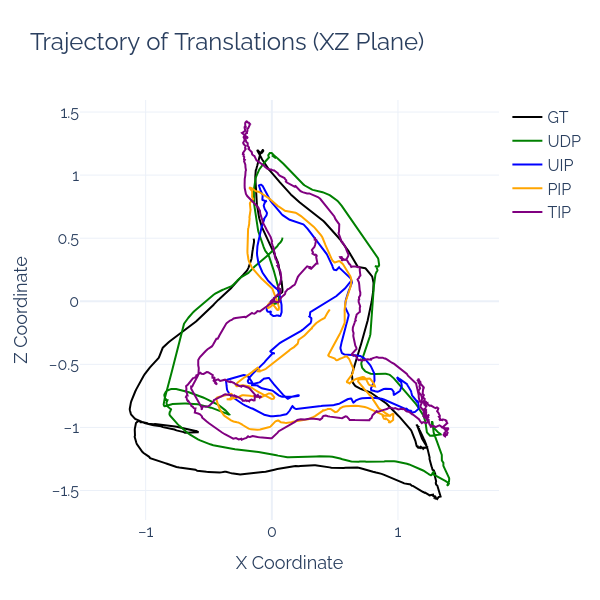}
    \caption{An example root trajectory in xz-plane. PIP and UIP tend to underestimate translation. \projname follows the trajectory closer. TIP has a higher level of jitter.}
    \label{fig:translation}
\end{figure}

\begin{figure}[ht]
    \centering
    \includegraphics[width=1\linewidth]{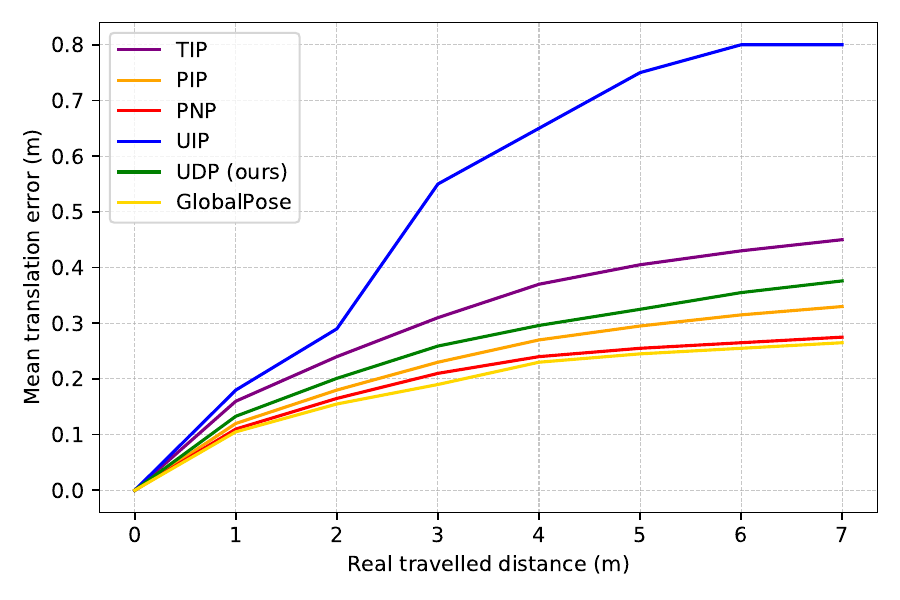}
    \caption{The cumulative translation error over segment of $d$ meters on TotalCaptures. All models, but UIP, perform similarly in translation quality.}
    \label{fig:translation_error_total_capture}
\end{figure}

\projname's primary contribution lies in improving the pose prediction by modeling geometric constraints using inter-sensor measurements.
As demonstrated in the main paper, the Spatial Layout Module and UWB-Diffusion Guidance significantly improve local pose estimation, particularly for arm and leg placement.
In this section, we describe the modeling of translation prediction and the evaluation of \projname's translation prediction.

\noindent \textbf{Translation Representation.} Unlike previous methods \cite{PIP, yi2024physical, TIP, UIP, diffusionPoser} that model translation implicitly by predicting frame-to-frame velocity, \projname outputs absolute translation predictions for individual window, each originating at the origin.
To form a continuous, long-term sequence, these windows are concatenated such that each window begins at the final position of the preceding one.
Velocity-based approaches require numerical integration to obtain global positions, a process where small inaccuracies accumulate over time.
By representing translation as an absolute position directly, we aim to mitigate this integration drift.

\noindent\textbf{Translation Evaluation.} We investigate translation prediction performance using the established cumulative translation error, which measures the drift from the ground truth trajectory after the subject has traveled a distance of $d$ m.
We report results on DanceDB and TotalCapture for all methods for which metrics are reported or could be reproduced directly.
Overall, \projname achieves translation accuracy that is competitive with existing approaches across all datasets. 
On DanceDB, it outperforms the baseline methods, see \cref{fig:translation_error_dance_db}. 
In contrast, on TotalCapture, methods that incorporate a physics optimizer, i.e., PIP, PNP, and GlobalPose, achieve lower drift, outperforming \projname by approximately 5 cm over 5 m (see \cref{fig:translation_error_total_capture}).

We investigate the varying results qualitatively.
On DanceDB, \projname generally yields more dynamic and accurate translations, whereas PIP and UIP tend to underestimate displacement, especially during twirls or intentional foot sliding.
This is visible in \cref{fig:translation}, where \projname follows the ground truth trajectory closely.

On TotalCapture, \projname exhibits drift when the translation direction is estimated incorrectly.
While the magnitude of fast, large-scale movements is tracked well, small directional errors can occasionally accumulate into larger drift.

We attribute these findings to two primary factors.
First, \projname significantly improves foot and leg placement. 
Since human translation is intrinsically linked to gait, improved estimation of stride length and foot positioning reduces translation error, as observed on DanceDB.
Second, physics optimized baselines benefit from explicit contact and dynamics constraints, which aid in tracking translation, as seen on TotalCapture. 
However, strict contact priors, such as zero-velocity assumptions, may degrade performance when the motion violates these constraints, such as during the intentional sliding maneuvers in DanceDB.

Overall,  \projname demonstrates strong performance in local pose estimation due to the incorporation of on-body inter-sensor constraints, though it yields mixed results regarding global translation. 
Future work could investigate how pairing physics optimization with \projname, or the incorporation of external UWB anchors, can further improve translation accuracy.

\subsubsection{Ablation: Absolute Translation Prediction}
Existing methods such as UIP, PIP, TIP, and DiffusionPoser predict between-frame translation, i.e., velocity.
In contrast, we employ an absolute translation representation, normalized to the origin.
This representation improves not only the translation results but also marginally enhances local pose estimation. 
Experimentally, switching to a velocity-based representation increases translation drift by 16\,cm over a 5\,m distance. 
As shown in \cref{tab:ablation_full}, \projname yields superior performance in both local pose and translation when predicting positions directly, highlighting the benefit of this approach.

% \begin{table}[ht]
%     \centering
%     \setlength{\tabcolsep}{3pt}
%     \caption{Ablation (kinematics losses: velocity \& FK) on TotalCapture.}
%     \scalebox{0.9}{
%     \begin{tabular}{lccc}
%     \toprule
%          Metric & SIP (°) $\downarrow$ & GAE (°) $\downarrow$ & JPE (cm) $\downarrow$ \\
%          \midrule
%          \projname w/ velocity & 9.30 & 10.35 & 3.95 \\
%          \projname w/ FK loss & 9.01 & 10.29 & 3.89 \\
%          \bottomrule
%           \projname w/o UWB & 14.20 & 12.64 & 6.44 \\
%     \end{tabular}
%     }
%     \label{tab:ablation_kinematics}
% \end{table}

\end{document}